\documentclass{article}
\usepackage[utf8]{inputenc}
\usepackage[T1]{fontenc}
\usepackage[final]{colm2026_conference}
\usepackage{microtype}
\usepackage{hyperref}
\usepackage{url}
\usepackage{booktabs}
\usepackage{tcolorbox}
\usepackage{multirow}
\usepackage{xcolor}
\usepackage{enumitem}
\usepackage{amsmath}
\usepackage{graphicx}
\usepackage{placeins}
\usepackage{wrapfig}

\usepackage{caption}
\makeatletter
\def\section{\@startsection{section}{1}{\z@}{-1.4ex plus -0.4ex minus -.2ex}%
    {0.9ex plus 0.2ex minus 0.1ex}{\large\bf\raggedright}}
\def\subsection{\@startsection{subsection}{2}{\z@}{-1.2ex plus -0.3ex minus -.2ex}%
    {0.5ex plus .1ex}{\normalsize\bf\raggedright}}
\def\subsubsection{\@startsection{subsubsection}{3}{\z@}{-1.0ex plus -0.3ex minus -.2ex}%
    {0.4ex plus .1ex}{\normalsize\bf\itshape\raggedright}}
\makeatother

\usepackage{tikz}
\tcbuselibrary{skins}
\usetikzlibrary{calc,positioning}
\definecolor{p0gray}{HTML}{6B7280}
\definecolor{p1blue}{HTML}{3B82F6}
\definecolor{p2green}{HTML}{16A34A}
\definecolor{p2violet}{HTML}{8B5CF6}
\definecolor{p3orange}{HTML}{EA580C}
\definecolor{outerframe}{HTML}{4B5563}

\graphicspath{{figures/}}
\definecolor{darkblue}{rgb}{0, 0, 0.5}
\hypersetup{colorlinks=true, citecolor=darkblue, linkcolor=darkblue, urlcolor=darkblue}
\title{When Audio-Language Models Fail to Leverage Multimodal Context for Dysarthric Speech Recognition}
\author{Pehu\'en Moure$^{1}$\thanks{Equal contribution.}, Niclas Pokel$^{1}$\footnotemark[1], Bilal Bounajma$^{2}$,\\Yingqiang Gao$^{3}$,
Roman Boehringer$^{1}$, Longbiao Cheng$^{1}$, Shih-Chii Liu$^{1}$ \\[4pt]
$^{1}$Institute of Neuroinformatics, University of Zurich and ETH Zurich, Switzerland \\
$^{2}$Department of Mathematics, ETH Zurich, Switzerland \\
$^{3}$Department of Computational Linguistics, University of Zurich, Switzerland \\
\texttt{\{pehuen,npokel,roman,shih\}@ini.ethz.ch, bbounajma@student.ethz.ch}\\
\texttt{longbiao@ini.uzh.ch, yingqiang.gao@cl.uzh.ch}
}
\definecolor{improve}{HTML}{22C55E}
\definecolor{degrade}{HTML}{EF4444}
\definecolor{neutral}{HTML}{9CA3AF}
\newcommand{\imp}[1]{\textcolor{improve}{\textbf{#1}}}
\newcommand{\degrad}[1]{\textcolor{degrade}{\textbf{#1}}}
\newcommand{\neu}[1]{\textcolor{neutral}{#1}}
\newcommand{\sig}[1]{\textsuperscript{#1}}
\begin{document}
\maketitle
\begin{abstract}
Automatic speech recognition (ASR) systems remain brittle on dysarthric and other atypical speech. Recent audio-language models raise the possibility of improving performance by conditioning on additional clinical context at inference time, but it is unclear whether these models can make use of such information. We introduce an evaluation protocol built on the Speech Accessibility Project (SAP) dataset that tests whether diagnosis labels, clinician-derived speech ratings, and progressively richer clinical descriptions improve transcription accuracy for dysarthric speech. Across matched comparisons on ten models, we find that current models do not meaningfully use this context: diagnosis-informed and clinically detailed prompts yield negligible improvements and often degrade word error rate, and counterfactual controls reveal that the models respond to the presence of prompt text rather than to its content. We complement the prompting analysis with context-dependent fine-tuning, showing that LoRA adaptation with a mixture of clinical prompt formats achieves a WER of 0.066, a 52\% relative reduction over the frozen baseline, while preserving performance when context is unavailable. The residual benefit of context concentrates on the hardest utterances, and a difficulty-aware gate that applies clinical context only where it helps lowers WER by about 10\% relative over the audio-only baseline on both architectures. These results clarify where current models fall short and provide a testbed for measuring progress toward more inclusive ASR.
\end{abstract}
\section{Introduction}
\label{sec:intro}
Modern ASR systems have achieved near-human performance on standard benchmarks~\citep{radford2022robustspeechrecognitionlargescale, Zhang_2022}, but these gains are unevenly distributed. Speakers with dysarthria, a motor speech disorder arising from neurological conditions such as cerebral palsy (CP), amyotrophic lateral sclerosis (ALS), Parkinson's disease, and stroke, experience dramatically higher word error rates (WER) \citep{tobin2024asr_disordered_speech}. This gap renders voice-based interfaces such as voice assistants, augmentative and alternative communication (AAC\footnote{\url{https://www.asha.org/public/speech/disorders/aac/}}) devices, and clinical documentation tools effectively unusable for millions of people, excluding them from an increasingly voice-first digital infrastructure~\citep{hoffman2014nhis, jaddoh2025overcomingspeechbarriers}.

\begin{figure}[tbp]
\centering
\includegraphics[width=0.93\textwidth]{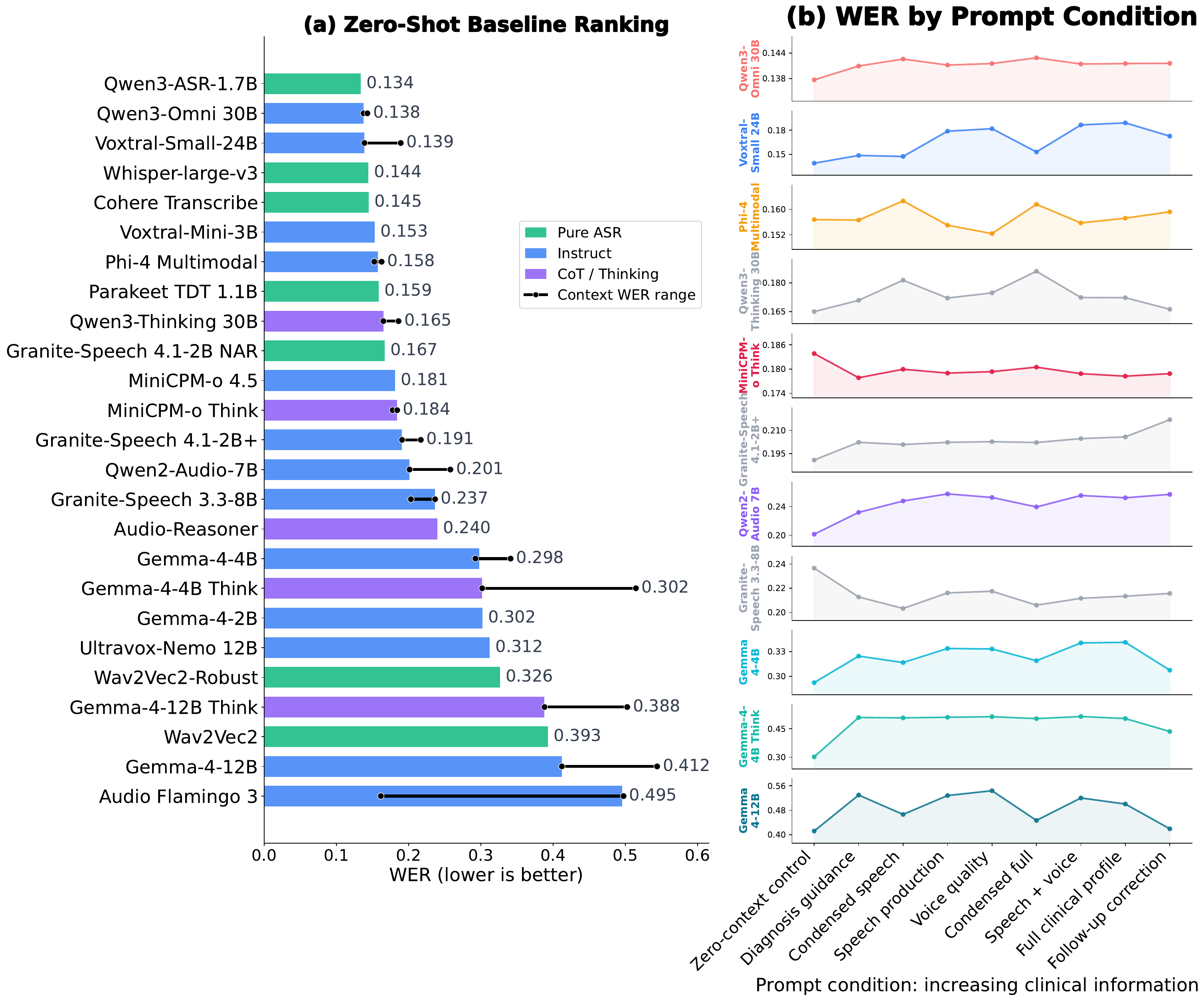}
\caption{(a)~ASR baseline ranking on dysarthric speech (zero-shot WER). Bars are colored by model type; black markers show the WER range across prompt conditions for promptable models. (b)~Per-model WER across prompt conditions ($N = 11{,}218$ matched samples). Each facet shows one model on its own scale. Frozen models either remain flat or degrade as clinical context increases. AF3's apparent gain with context reflects suppressed hallucination rather than genuine context use (Section~\ref{sec:results_conditioning}).}
\label{fig:baselines_and_conditioning}
\end{figure}

A new opportunity has emerged with audio-native multimodal LLMs such as Gemini~2.0~\citep{geminiteam2025geminifamilyhighlycapable} and GPT-4o~\citep{openai2024gpt4ocard}, which process raw audio directly alongside natural-language instructions. These models exhibit strong in-context learning on natural language processing (NLP) tasks \citep{dong2024surveyincontextlearning}; the open question is whether this capability extends to audio understanding of atypical speech. Clinical metadata about a speaker's condition (diagnosis, severity, articulatory characteristics) is routinely available in medical and therapeutic settings \citep{morris2024documentationdisability} and could, in principle, serve as a natural-language ``adapter'' at inference time without model retraining.

We make the following contributions. Working on the Speech Accessibility Project (SAP, \cite{zheng2025interspeech2025speechaccessibility}) dataset, first, we introduce a systematic evaluation protocol that isolates the effect of different clinical context types on dysarthric ASR, spanning pathology-informed prompting with progressively richer clinical profiles, follow-up correction, and counterfactual context controls (wrong clinical content, length-matched random text, and instruction-only filler). Second, we conduct the first (to the best of our knowledge) broad assessment of ten audio-language models under controlled clinical prompting conditions on dysarthric speech, within a zero-shot comparison of 25 recent ASR and audio-language systems, showing that frozen models consistently fail to exploit clinical context, with three distinct failure modes (robust, degrading, and format-dependent). Third, we show on two model families (Voxtral-Small-24B and Gemma-4-E2B) that context-dependent fine-tuning (LoRA adaptation \citep{hu2021lora} on a mixture of clinical prompt formats) removes the prompt-induced degradation and achieves the lowest WER of any system we evaluate while preserving performance when context is unavailable; counterfactual controls reveal that the fine-tuned models respond to the presence of clinical context rather than to its graded content. This indicates that the poor performance of multimodal models is due to the absence of similar data samples in their training data rather than some fundamental limitation. Fourth, we show that the residual benefit of context concentrates on the hardest utterances and formulate a difficulty-aware gate that applies clinical context only where it helps, lowering WER by about 10\% relative over the audio-only baseline on both architectures. We publicly release our evaluation code, prompt templates, and evaluation pipeline.

\section{Related Work}
\label{sec:related}
\paragraph{ASR for dysarthric and atypical speech.}
Traditional approaches to dysarthric ASR include speaker-dependent GMM-HMM~\citep{rabiner1989tutorial} and DNN-HMM~\citep{bhat2016dysarthric} systems, data augmentation strategies using tempo and pitch perturbation \citep{BHAT2025109954}, and TTS-based synthesis of dysarthric speech for training data expansion \citep{hermann2023fewshot}. More recent work has focused on fine-tuning self-supervised models such as wav2vec~2.0 \citep{modelwav2vec2robust} or Whisper \citep{wagner2025personalizedfinetuningcontrollablesynthetic} on dysarthric corpora, achieving substantial improvements over unadapted baselines. Recent work has also explored data-efficient Whisper-based personalization under limited-data conditions by constructing semantically structured training material from small amounts of user-specific speech \citep{pokel2025semanticrechaining}. In parallel, parameter-efficient adaptation methods such as low-rank and Bayesian low-rank \citep{pokel2025vilora} fine-tuning have also been investigated for impaired speech recognition. A key limitation shared by all these approaches is the requirement for training or fine-tuning on data from the target speaker population, which is normally expensive to collect.

Closest to our setting are systems built on SAP data itself. The winner of the Interspeech 2025 SAP Challenge fully fine-tunes Parakeet-TDT and reaches 5.97\%/8.11\% WER on the challenge Test1/Test2 splits \citep{takahashi2025parakeet}, and a concurrent proprietary system based on Gemini~2.5~Flash reaches about 5\% WER through few-shot same-speaker in-context personalization \citep{agarwal2025personalizer}. Both use full fine-tuning or speaker personalization and are measured on the challenge private test splits, not on our rated subset, so they serve as a complementary upper bound rather than a matched comparison. Anchored to a frozen Whisper baseline of comparable difficulty across splits (14--18\% WER), our parameter-efficient context-dependent LoRA (Section~\ref{sec:results_supervised}) lands within one to two points of these systems at far lower adaptation cost (Appendix~\ref{app:positioning}). The personalization result independently supports same-speaker audio-based in-context learning, a direction we flag as future work.

\paragraph{Multimodal LLMs with audio understanding.}
The Gemini model family \citep{geminiteam2025geminifamilyhighlycapable} and GPT-4o \citep{openai2024gpt4ocard} natively accept audio input alongside text, enabling audio understanding without a separate ASR front-end. Open-weight alternatives include Qwen2-Audio \citep{modelqwen2} and Qwen3-Omni \citep{modelqwen3omni}. Evaluations of these models have focused predominantly on standard speech recognition benchmarks, speaker identification, and audio captioning. To our knowledge, no systematic evaluation exists for their performance on pathological or atypical speech under controlled prompting conditions.

\paragraph{In-context learning for speech and audio.}
In-context learning is well-studied in the text domain \citep{dong2024surveyincontextlearning} and increasingly explored for vision tasks \citep{zhou2024visualincontextlearninglarge}. For speech and audio, in-context learning remains underexplored, with most work focusing on few-shot prompting for text-based NLP tasks rather than direct audio conditioning. A recent work \citep{alsayegh2025zeroshotrecognitiondysarthricspeech} evaluated commercial ASR systems and multimodal large language models on dysarthric speech using the TORGO corpus \citep{rudzicz2012torgo}, showing strong performance degradation with increasing severity and only limited benefits from simple prompting strategies. However, this line of work remains focused on zero-shot transcription and does not provide a systematic analysis of how different forms of clinical context affect recognition.

\section{Evaluation Protocol}
\label{sec:method}
We present our evaluation protocol, consisting of a systematic clinical context conditioning assessment pipeline and a supervised fine-tuning adaptation study. The aim is to measure whether (and how) audio-language models can leverage increasing levels of clinical contextual information to improve dysarthric speech recognition (Figure~\ref{fig:overview}).

\begin{figure}[tbp]
\centering
\resizebox{\linewidth}{!}{%
\begin{tikzpicture}[
  x=1cm, y=1cm,
  blk/.style={draw=outerframe, fill=gray!8, rounded corners=2.5pt, line width=0.9pt},
  lbl/.style={font=\sffamily\scriptsize},
  tin/.style={font=\sffamily\tiny},
  arr/.style={-stealth, line width=0.9pt, draw=outerframe}
]
\node[draw=outerframe, rounded corners=2pt, line width=0.8pt, fill=gray!4,
      text width=4.15cm, align=left, inner sep=4pt, anchor=north west] (p0) at (0, 4.55)
  {{\sffamily\bfseries\tiny\textcolor{outerframe}{P0}}\;{\sffamily\tiny audio only}\\[1pt]
   {\ttfamily\tiny Transcribe the audio in English.}};
\node[draw=p1blue, rounded corners=2pt, line width=0.8pt, fill=p1blue!4,
      text width=4.15cm, align=left, inner sep=4pt, below=2.5mm of p0.south west, anchor=north west] (p1)
  {{\sffamily\bfseries\tiny\textcolor{p1blue}{P1}}\;{\sffamily\tiny + diagnosis}\\[1pt]
   {\ttfamily\tiny The speaker has Cerebral Palsy, which often causes [...]}};
\node[draw=p2violet, rounded corners=2pt, line width=0.8pt, fill=p2violet!4,
      text width=4.15cm, align=left, inner sep=4pt, below=2.5mm of p1.south west, anchor=north west] (p2)
  {{\sffamily\bfseries\tiny\textcolor{p2violet}{P2}}\;{\sffamily\tiny + clinical profile}\\[1pt]
   {\ttfamily\tiny [...P1...] Clinical profile (1--7): imprecise consonants 4/7 $\cdot$ [...]}};
\node[font=\sffamily\bfseries\tiny, anchor=south west, text=outerframe] at (0, 4.65) {Prompt (varies)};
\node[font=\sffamily\bfseries\tiny, anchor=south west, text=outerframe] at (4.7, 4.65) {Audio (fixed)};
\foreach [count=\i] \h in {0.12,0.5,0.2,0.78,0.3,0.9,0.24,0.55,0.14,0.7,0.4,0.82,0.18,0.48,0.12}
  \draw[line width=1.0pt, line cap=round, draw=p2green]
    ({4.6+\i*0.105}, {4.25-\h/2}) -- ({4.6+\i*0.105}, {4.25+\h/2});
\draw[blk] (6.8,0.95) rectangle (12.05,4.95);
\node[lbl, align=center, anchor=north] at (9.425,4.89)
  {\bfseries Audio-Language Model\\{\sffamily\tiny\textcolor{p0gray}{(frozen)}}};
\draw[draw=outerframe, fill=p2green!8, rounded corners=2.5pt, line width=0.9pt]
  (7.0,3.2) -- (8.6,3.4) -- (8.6,4.1) -- (7.0,4.3) -- cycle;
\node[font=\sffamily\bfseries\tiny, align=center] at (7.8,3.75) {Audio\\Embeddings};
\draw[draw=outerframe, fill=p1blue!8, rounded corners=2.5pt, line width=0.9pt]
  (7.0,1.25) -- (8.6,1.45) -- (8.6,2.15) -- (7.0,2.35) -- cycle;
\node[font=\sffamily\bfseries\tiny, align=center] at (7.8,1.8) {Text\\Embeddings};
\foreach \i in {0,1,2} \draw[draw=p1blue!60, fill=p1blue!22, line width=0.4pt]
  ({9.7+\i*0.32},1.35) rectangle ({9.85+\i*0.32},3.85);
\node[lbl, text=p0gray] at (10.85,2.6) {$\cdots$};
\foreach \i in {0,1} \draw[draw=p1blue!60, fill=p1blue!22, line width=0.4pt]
  ({11.15+\i*0.32},1.35) rectangle ({11.3+\i*0.32},3.85);
\node[tin, text=p0gray, anchor=north] at (9.78,1.3) {L$_{1}$};
\node[tin, text=p0gray, anchor=north] at (11.55,1.3) {L$_{N}$};
\node[draw=p3orange, rounded corners=2pt, line width=0.8pt, fill=p3orange!8,
      inner sep=4pt, anchor=west] (out) at (12.45, 2.9)
  {\ttfamily\tiny turn off the TV};
\node[font=\sffamily\bfseries\tiny, anchor=south, text=outerframe] at ($(out.north)+(0,0.08)$) {Transcript};
\node[tin, align=center, anchor=north, text=outerframe] at ($(out.south)+(0,-0.35)$)
  {WER $\downarrow$\\vs.\ reference};
\draw[arr, draw=p2green] (6.28,4.25) to[out=0, in=150] (6.97,3.9);
\draw[arr] (p0.east) to[out=-40, in=160] (6.97,1.95);
\draw[arr, draw=p1blue] (p1.east) to[out=0, in=180] (6.97,1.8);
\draw[arr, draw=p2violet] (p2.east) to[out=0, in=195] (6.97,1.65);
\draw[arr, draw=p2green] (8.6,3.75) to[out=0, in=150] (9.62,3.1);
\draw[arr] (8.6,1.8) to[out=0, in=210] (9.62,2.1);
\draw[arr] (12.05,2.9) -- (12.42,2.9);
\draw[arr] ($(out.south)+(0,-0.05)$) -- ($(out.south)+(0,-0.3)$);
\end{tikzpicture}%
}
\caption{Overview of the evaluation protocol. The same frozen audio-language model receives identical audio while the text prompt varies from a bare transcription instruction (P0) to progressively richer clinical context (P1, P2; Section~\ref{sec:prompt_conditions}); outputs are scored by WER against reference transcripts. Counterfactual controls (Section~\ref{sec:results_counterfactual}) replace the clinical block with wrong, random, or instruction-only text of matched length.}
\label{fig:overview}
\end{figure}

We use standard ASR metrics (WER, CER) and SemScore~\citep{phukon2025aligningasrevaluationhuman}, detailed in the Appendix~\ref{app:metrics}, as well as the conventions from the SAP dataset \citep{zheng2025interspeech2025speechaccessibility}: we clip per-sample WER at 1.0, use dual-reference minimum scoring, and apply Whisper's \texttt{EnglishTextNormalizer}~\citep{radford2022robustspeechrecognitionlargescale}.

\subsection{Prompt Conditions}
\label{sec:prompt_conditions}
We define a hierarchy of prompt conditions with progressively richer clinical context, from a zero-context control (P0) through diagnosis guidance (P1) to full clinical profiles (P2); see Table~\ref{tab:conditions} in Appendix~\ref{app:prompts} for the full prompts.
The clinical ratings are drawn from two complementary profiles: the \emph{speech production profile} captures articulatory characteristics such as imprecise consonants, slow rate, and hypernasality, and the \emph{voice quality profile} captures phonation and vocal-control cues such as breathiness, tremor, and monopitch (full dimension lists in Appendix~\ref{app:prompts}).

Beyond this hierarchy, three counterfactual control conditions preserve the structure of the full clinical profile while manipulating its content: a \emph{wrong-context} condition in which the diagnosis and per-sample ratings are replaced with those of a different speaker (cyclic rotation across etiologies), a \emph{length-matched random text} condition in which the clinical block is replaced by unrelated text of matched length, and a \emph{length-matched instruction-only} condition in which it is replaced by generic transcription instructions (templates in Appendix~\ref{app:prompts}).

\subsection{Models Under Evaluation}
\label{sec:models}
We evaluate ten audio-language models under the full prompt-condition battery: Audio Flamingo~3 \citep{modelaudioflamingo3}, MiniCPM-o~4.5 Think \citep{modelminicpm}, Qwen2-Audio-7B \citep{modelqwen2}, Qwen3-Omni-30B \citep{modelqwen3omni}, Qwen3-Thinking-30B \citep{modelqwen3omni}, Phi-4 Multimodal \citep{modelphi4}, Voxtral-Small-24B \citep{modelvoxtralsmall}, Gemma-4-4B \citep{modelgemma}, Gemma-4-4B Think \citep{modelgemma}, and Gemma-4-12B \citep{modelgemma} (a unified, encoder-free multimodal model).
Five further audio-language models \citep{modelvoxtralsmall, modelultravox, modelgemma, modelminicpm, modelaudioreasoner} are included in the zero-shot ranking only (Section~\ref{sec:results_baselines}, Table~\ref{tab:models}), as compute constraints precluded the full prompt grid.
As baselines, we report nine ASR systems \citep{radford2022robustspeechrecognitionlargescale, modelQwen3ASR, modelparakeet, modelcohere, modelgranitespeech, modelwav2vec2robust, modelwav2vec2960h}, including Whisper large-v3, Parakeet-TDT-1.1B, and two legacy Wav2Vec2 models (complete list in Table~\ref{tab:models}). Supervised adaptation (Section~\ref{sec:results_supervised}) is conducted on Voxtral-Small-24B and replicated on a second model family, Gemma-4-E2B (listed as Gemma-4-2B in Table~\ref{tab:models}).

\subsection{Dataset}
\label{sec:dataset}
We conduct our clinical context conditioning evaluation on the Speech Accessibility Project (SAP, \citet{zheng2025interspeech2025speechaccessibility}) dataset, the largest available dysarthric speech corpus. SAP contains recordings from 959~speakers across five etiologies (Parkinson's disease, ALS, cerebral palsy, Down syndrome, and stroke).

A subset of samples contains clinician-provided ratings across perceptual dimensions (e.g., imprecise consonants, breathy voice, monopitch), which form the basis for our clinical prompt conditions. The rated subset used for prompt-condition comparisons comprises $N = 11{,}218$ matched samples, drawn from four etiologies (Parkinson's disease, ALS, cerebral palsy, and Down syndrome); stroke speakers are absent from this subset.

\section{Results}
\label{sec:results}

\subsection{Zero-Shot Baselines}
\label{sec:results_baselines}
Figure~\ref{fig:baselines_and_conditioning}(a) and Table~\ref{tab:baselines} rank all models by zero-shot WER on the SAP rated subset. Qwen3-ASR-1.7B achieves the lowest WER (0.134), followed by Qwen3-Omni-30B (0.138), Voxtral-Small-24B (0.139), Whisper large-v3 (0.144), and Cohere Transcribe (0.145). The conventional speech recognition system Wav2Vec2 (0.393) lags substantially behind (full results in Table~\ref{tab:zero_shot_results}).

\subsection{Clinical Context Conditioning of Frozen Models}
\label{sec:results_conditioning}
Across all ten models evaluated on the full set of prompt conditions, adding clinical context to the prompt either has a negligible effect or actively degrades WER. Figure~\ref{fig:baselines_and_conditioning}(b) shows this trend (with the exception of Audio Flamingo~3, as discussed below). The full condition matrix and paired comparisons are in Tables~\ref{tab:conditioning} and~\ref{tab:paired_headline} (Appendix~\ref{app:main_tables}).\footnote{All reported $p$-values use Benjamini--Hochberg false-discovery-rate (FDR) correction across the 50 model~$\times$~condition comparisons.}

\begin{figure}[tbp]
\centering
\includegraphics[width=\textwidth]{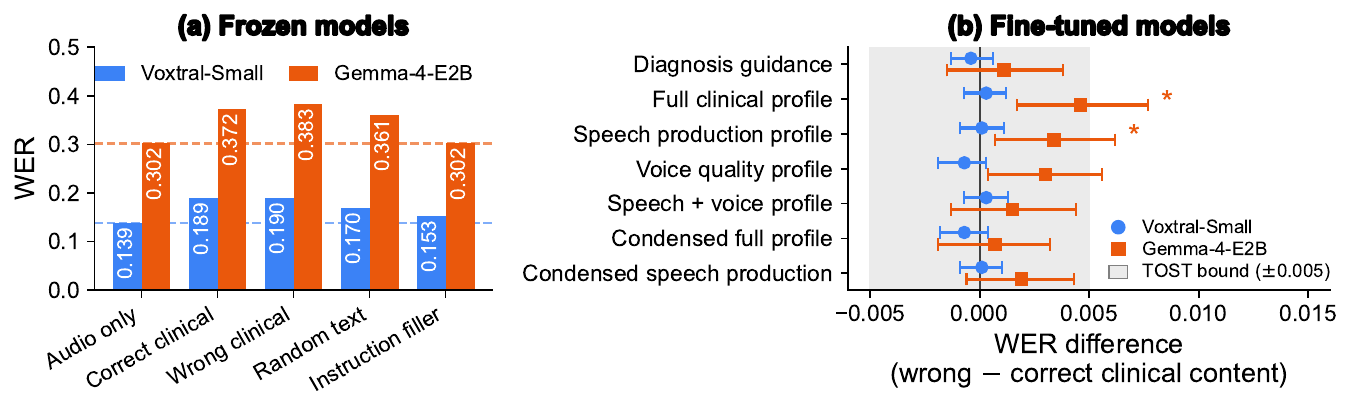}
\caption{Counterfactual context controls on two architectures. (a)~Frozen WER by prompt condition; dashed lines mark the audio-only baselines. (b)~WER difference between wrong and correct clinical content for the fine-tuned context-dependent systems. Voxtral-Small is equivalent throughout, and Gemma-4-E2B shows a small correct-content gain at two conditions ($\ast$).}
\label{fig:counterfactual}
\end{figure}

The results reveal three failure modes. \textbf{Robust models} (Phi-4, Qwen3-Omni, MiniCPM-o Think, Qwen3-Thinking) show negligible WER change under clinical context ($-0.006$ to $+0.007$). \textbf{Degrading models} (Voxtral-Small, Qwen2-Audio, Gemma-4-4B, Gemma-4-4B Think, Gemma-4-12B) show significant WER increases under clinical prompting, with Gemma-4-4B Think exhibiting catastrophic failure ($\Delta$ WER $+0.202$, 42.7\% of samples degraded) as the model's chain-of-thought process compounds with prompt length and displaces the transcription itself (Section~\ref{sec:analysis}).
\textbf{Format-dependent models} (Audio Flamingo~3) appear to improve markedly under prompting, but AF3's unprompted WER of 0.495 reflects non-transcription output (audio descriptions, commentary) in the absence of a task instruction; the structured prompt merely restores baseline transcription behavior ($\approx$0.162), and AF3 is nearly insensitive to the prompt's clinical content (0.162--0.164).

Error decomposition (Table~\ref{tab:error_decomp}, Appendix~\ref{app:main_tables}) gives a mechanistic account, with increased insertions for Voxtral-Small and Gemma-4-12B, whereas Gemma-4-4B Think degrades through truncation instead (category breakdown in Table~\ref{tab:category_strat}; mechanisms in Section~\ref{sec:analysis}).

\subsection{Counterfactual Context Controls}
\label{sec:results_counterfactual}
The degradation documented above could stem from models reacting to the clinical \emph{content} of the prompt or merely to the \emph{presence} of additional prompt text. The control conditions (Section~\ref{sec:prompt_conditions}) separate these explanations on two architectures, Voxtral-Small and the smaller Gemma-4-E2B (Figure~\ref{fig:counterfactual}a).
On both models, wrong clinical content is no better than correct content, length-matched random text degrades almost as much, and length-matched instruction-only filler lies closest to the audio-only baseline, ruling out prompt length as the cause. The frozen degradation therefore tracks the presence of non-instructional prompt text rather than its clinical accuracy, and the effect replicates across model families (Figure~\ref{fig:counterfactual}a; Table~\ref{tab:frozen_controls}, Appendix~\ref{app:counterfactual_tables}; fine-tuned controls in Section~\ref{sec:results_supervised}).

\paragraph{Severity and etiology stratification.}\label{sec:results_stratified}
Figure~\ref{fig:severity_wer} (Appendix~\ref{app:additional}) shows how baseline WER increases sharply with clinical severity across all models, with the steepest degradation for cerebral palsy and Down syndrome beyond severity level~2. The degradation pattern from prompting is broadly consistent across severity levels and etiologies. No subgroup shows reliable improvement under clinical prompting for frozen models. Full numerical breakdowns are in Appendix~\ref{app:main_tables} Tables~\ref{tab:severity_strat} and~\ref{tab:etiology_strat}.

\paragraph{Iterative correction.} Providing a prior-pass transcript alongside the full clinical profile does not systematically improve WER over single-pass prompting; Voxtral-Small and Qwen2-Audio remain above their zero-context baselines (Appendix~\ref{app:additional}).

\subsection{Supervised Adaptation}
\label{sec:results_supervised}
\begin{wrapfigure}{r}{0.52\textwidth}
\centering
\includegraphics[width=\linewidth]{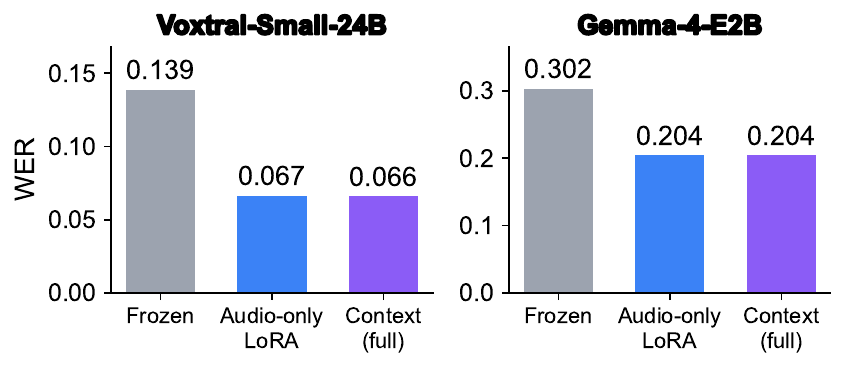}
\caption{Fine-tuning comparison on both architectures (5-fold cross-validation, 437 speakers). LoRA halves WER on Voxtral-Small (0.139 $\to$ 0.067) and reduces Gemma-4-E2B from 0.302 to 0.204; the context-dependent systems match audio-only fine-tuning. Voxtral-Small values in Table~\ref{tab:supervised_summary}.}
\label{fig:supervised_results}
\end{wrapfigure}
Given that frozen multimodal models uniformly fail to exploit clinical context, we investigate whether supervised fine-tuning can teach a model to use it. We fine-tune Voxtral-Small-24B with LoRA using 5-fold speaker-disjoint cross-validation across 437 unique speakers (87--88 held out per fold), yielding 11{,}218 utterances pooled out-of-fold. We train two systems. The \textbf{audio-only} system is fine-tuned with audio input only and no clinical prompt. The \textbf{context-dependent} system is fine-tuned on a mixture of audio-only, speech~+~voice profile, and condensed full profile samples, exposing the model to varied context formats during training (see Appendix~\ref{app:finetuning} for full training details); all results are pooled out-of-fold predictions from the last checkpoint per fold.

The audio-only system reduces WER from 0.139 (frozen) to 0.067, a 52\% relative reduction (Figure~\ref{fig:supervised_results}). The context-dependent system achieves the best overall WER of 0.066 when evaluated with the full clinical profile, a modest 0.8\% relative gain over audio-only that does not reach significance at the speaker level ($p = 0.55$); the same weights produce WER 0.067 at zero-context ($p = 0.31$; Table~\ref{tab:supervised_paired}). Notably, providing context incurs no penalty. The context-dependent model without context (0.067) performs comparably to audio-only (0.067, $p = 0.93$), indicating that context-dependent fine-tuning does not sacrifice audio-only quality.

\paragraph{Replication on a second model family.} To test whether these findings are specific to Voxtral-Small, we replicate the fine-tuning protocol on Gemma-4-E2B, the weakest promptable model family in our frozen evaluation (training details in Appendix~\ref{app:finetuning_gemma}). The pattern reproduces: audio-only LoRA reduces WER from 0.302 (frozen zero-context) to 0.204, the context-dependent system evaluated with the full clinical profile also reaches 0.204, and the frozen-model degradation under clinical prompting ($+0.070$ at the full profile) disappears entirely.

\paragraph{Counterfactual controls on the fine-tuned systems.} Does the context-dependent system read the clinical content, or only register its presence? We evaluate the same fine-tuned weights with correct versus wrong clinical profiles (wrong-context condition, Section~\ref{sec:prompt_conditions}) on the same utterances and test equivalence directly with two one-sided tests (TOST) at a bound of $\pm 0.005$ WER (Appendix~\ref{app:stats}). On Voxtral-Small, correct and wrong clinical content are statistically equivalent at every clinical prompt condition (TOST $p < 0.001$; Figure~\ref{fig:counterfactual}b), and this holds across three further training variants, including a curriculum-trained one (audio adaptation followed by context-only refinement). Gemma-4-E2B is informative in a different way: it shows a small but significant correct-over-wrong gain at the full clinical profile and speech production profile conditions (correct lower by about 0.004 WER, Wilcoxon $p \le 0.005$) and equivalence elsewhere (Table~\ref{tab:ft_counterfactual}, Appendix~\ref{app:counterfactual_tables}).

\paragraph{Subgroup analysis.}\label{sec:results_subgroups}
Table~\ref{tab:subgroups_main} (Appendix~\ref{app:additional}) compares audio-only $\to$ context-dependent (full clinical profile) by etiology and severity, showing relative WER change.

Down syndrome speakers show a statistically significant 7.1\% relative improvement ($p = 0.044$), and mild-severity speakers show a significant 2.7\% relative improvement ($p = 0.033$). ALS speakers show a near-significant 4.6\% improvement ($p = 0.066$). Cerebral palsy is the only etiology for which the context-dependent system slightly degrades (+0.006, not significant), consistent with the frozen-model findings.

Comparing these fine-tuning gains to the frozen-model results in Appendix~\ref{app:main_tables} Table~\ref{tab:etiology_strat} reveals a striking reversal. Under frozen-model clinical prompting (P0 $\to$ P2), the average across the ten evaluated models is a $\Delta$WER of $+$0.042 for Down syndrome and $+$0.029 for cerebral palsy speakers, with Down syndrome degrading on a majority of models and cerebral palsy on five of ten. After context-dependent fine-tuning, the same clinical information that degraded performance for Down syndrome speakers under a frozen model now yields the largest relative improvement ($-$7.1\%, Table~\ref{tab:subgroups_main}). The etiology ordering largely inverts: groups most harmed by frozen prompting are most helped by fine-tuning. This suggests that the clinical context contains useful signal for these populations that only becomes accessible once the model is trained to use it.
Section~\ref{sec:results_difficulty} and the analysis in Section~\ref{sec:context_bene} trace these differences to baseline acoustic difficulty.

\begin{figure}[tbp]
\centering
\includegraphics[width=\textwidth]{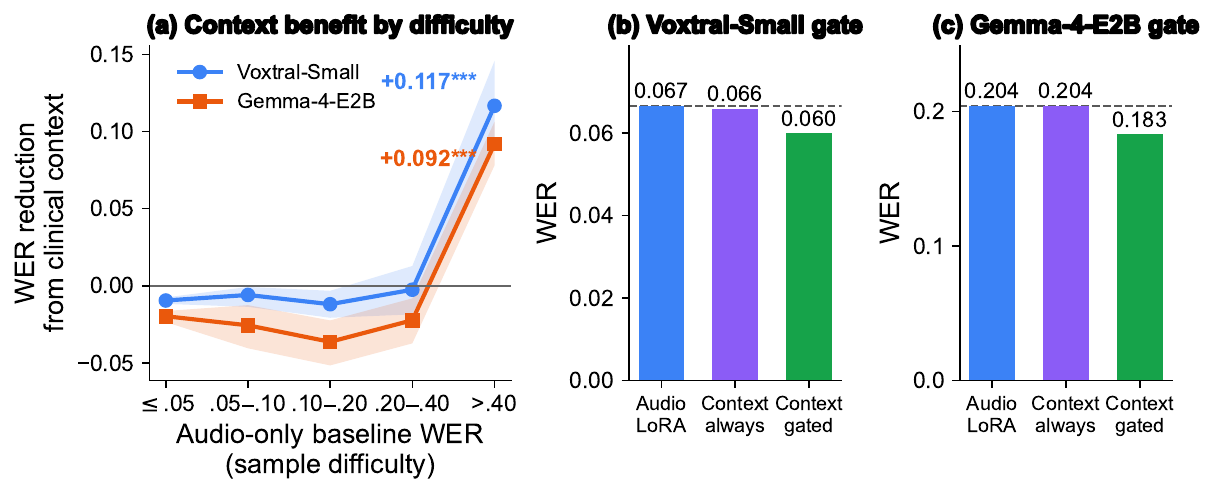}
\caption{Clinical context yields the largest gains on the most difficult utterances (5-fold cross-validation, both architectures). (a)~WER reduction from the full clinical profile as a function of audio-only sample difficulty; shaded bands are speaker-clustered 95\% intervals. The benefit concentrates on the hardest utterances, while easy utterances incur a small penalty. (b,~c)~Applying context only to hard utterances (difficulty-aware gate, oracle difficulty signal) lowers WER relative to the audio-only LoRA on both architectures; the dashed line marks the audio-only baseline.}
\label{fig:difficulty_gate}
\end{figure}

\subsection{A Difficulty-Aware Context Gate}
\label{sec:results_difficulty}
The aggregate parity between the audio-only and context-dependent systems hides a strong conditional effect. Holding the fine-tuned weights fixed, the benefit of the full clinical profile grows with sample difficulty, measured by the audio-only baseline WER (Spearman correlation, $p < 10^{-9}$; Figure~\ref{fig:difficulty_gate}a). On the hardest utterances, those the audio-only system transcribes at WER above 0.40, providing context lowers WER by 0.117 on Voxtral-Small and 0.092 on Gemma-4-E2B (Wilcoxon $p < 10^{-16}$, speaker-clustered intervals excluding zero). On easy utterances, context instead costs accuracy, slightly on Voxtral-Small and more on the weaker Gemma-4-E2B, so the two effects cancel in the aggregate.

The right policy is therefore to apply context only where it helps. Gating on sample difficulty, i.e.\ providing the clinical profile only for utterances the audio-only system finds hard, lowers overall WER from 0.0665 to 0.0599 on Voxtral-Small and from 0.2041 to 0.1833 on Gemma-4-E2B, a roughly 10\% relative reduction over the audio-only baseline on each architecture (Figure~\ref{fig:difficulty_gate}b,c). We compute the gate with an oracle difficulty signal (the audio-only baseline WER), so these numbers bound what a deployable gate can achieve; the no-reference difficulty signals we tested (system disagreement, severity ratings, audio features) do not yet recover this headroom, leaving a practical gating mechanism as an open problem. The result nevertheless reframes the role of clinical context: as expected, it is usable, and most valuable, precisely on the severely degraded speech where accessibility needs are greatest.

\section{Analysis and Discussion}
\label{sec:analysis}
\subsection{Insight on the Failure of Frozen Models}
The three failure modes identified in Section~\ref{sec:results_conditioning} suggest several distinct underlying mechanisms rather than a single explanation.

First, \textbf{weak conditioning on prompt content}. Robust models (Phi-4, Qwen3-Omni, MiniCPM-o Think, Qwen3-Thinking) may place much greater weight on the audio signal than on the semantic content of the text prompt, using the latter mainly to control output style or format. Audio Flamingo~3's near-zero sensitivity to prompt \emph{content} once output format is fixed is also consistent with this interpretation.

Second, \textbf{verbosity- and truncation-induced degradation}. For degrading models, longer clinical prompts shift the generation distribution away from faithful transcription. The error decomposition in Section~\ref{sec:results_conditioning} supports this finding: Voxtral-Small's insertion rate increases by $+$0.212 (verbose reformulation), while Gemma-4-4B Think's deletion rate increases by $+$0.174 (heavy truncation). The counterfactual controls (Section~\ref{sec:results_counterfactual}) refine the mechanism: the degradation is reproduced by length-matched random text but not by length-matched instruction-only filler, implicating non-instructional prompt content rather than clinical semantics or sheer prompt length.

Third, \textbf{distribution shift}. Clinical prompt formats are far from the training distribution of these models. Without exposure to the clinical-context-to-transcription mapping during training, the models cannot learn to condition on this information. 

The supervised results support the third hypothesis directly. Once Voxtral-Small or Gemma-4-E2B is fine-tuned on clinical prompts, the prompt-induced degradation disappears, and the same weights yield a difficulty-conditioned benefit from context (Section~\ref{sec:results_difficulty}). The counterfactual controls qualify what is learned: with the partial exception of the smaller Gemma model, the fine-tuned systems respond to the presence of clinical context rather than to its graded content. The limitation is not architectural; it is a matter of training exposure, though the exposure our recipe provides teaches models that context matters rather than what it says.

\begin{figure}[tbp]
\centering
\includegraphics[width=\textwidth]{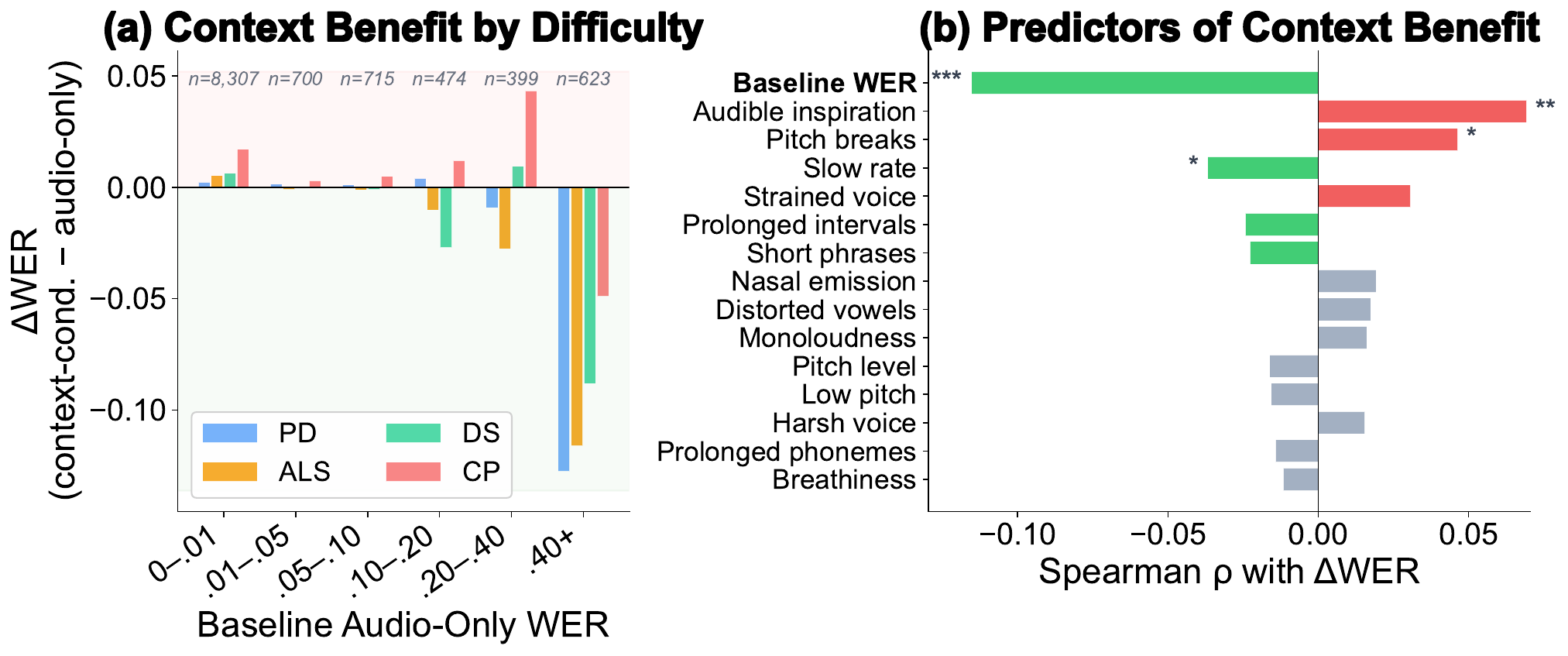}
\caption{Predictors of per-sample context benefit (audio-only $\to$ context-dependent, 5-fold cross-validation, $N = 11{,}218$). (a)~Mean $\Delta$WER by baseline difficulty and etiology. (b)~Spearman $\rho$ between each predictor and $\Delta$WER. Baseline WER dominates; individual clinical dimensions show near-zero correlation.}
\label{fig:context_benefit_predictors}
\end{figure}

\paragraph{When is context beneficial?}\label{sec:context_bene}
We regress per-sample $\Delta$WER (audio-only $\to$ context-dependent at the full clinical profile) against baseline difficulty, mean severity, etiology, and individual clinical rating dimensions (Figure~\ref{fig:context_benefit_predictors}). Baseline audio-only WER is the strongest predictor (Spearman $\rho = -0.12$, $p < 0.001$; Pearson $r = -0.24$), etiology and mean severity add little once difficulty is accounted for (additional $R^2 < 0.015$ each), and individual clinical rating dimensions show near-zero correlation with $\Delta$WER (the strongest is audible inspiration at $\rho = +0.07$, and most fall below $|\rho| = 0.03$), and all clinical prompt conditions improve WER by similar amounts (Table~\ref{tab:ctx_cond_grid}). The subgroup findings of Table~\ref{tab:subgroups_main} are therefore better read as a consequence of baseline difficulty distributions across etiologies rather than etiology-specific use of the clinical profiles.

\subsection{Hallucination, Verbosity and Refusal}
\label{sec:hallucination_analysis}
Clinical prompting also changes \emph{how} models fail. Defining a hallucination as an unclipped WER above 100\%, zero-context rates range from 2.6\% (Voxtral-Small) to 27.9\% (Audio Flamingo~3, driven by a preamble artifact that prompting removes, 27.9\% $\to$ 2.9\%), while Voxtral-Small rises to 8.3\%, Gemma-4-4B to 11.1\%, and Gemma-4-4B Think to 17.5\% under prompting; the remaining models vary by $\pm$1\,pp. Manual inspection identifies five recurring patterns: repetitive loops that echo the disfluencies described in the profile, verbose reformulation, language switching, refusal, and reasoning-chain leakage in CoT models. Only the first is directly attributable to clinical content, the model appearing to ``anticipate'' described disfluencies and reproduce them.

The unified Gemma-4-12B is the clearest instance of refusal. Clinical prompting raises its refusal rate from 2.4\% at zero-context to roughly 24\%, concentrated on short utterances such as digital-assistant commands (52\% refused): rather than transcribing, the model treats the spoken target as an instruction addressed to it and declines (``I cannot fulfill this request''). The refusals come with verbose non-transcription output (P90 output-to-reference length ratio $\approx$8--10$\times$), producing the largest insertion-rate increase of any model ($+$0.621, Table~\ref{tab:error_decomp}) and placing Gemma-4-12B among the degrading models despite competitive zero-context accuracy. Length ratios and fixed/induced decompositions are in Appendix~\ref{app:additional}.

\textbf{Word-level error decomposition.} Aligning the two systems word by word shows that context-dependent fine-tuning repairs 17\% of the consonant-word and 20\% of the vowel-word errors the audio-only system missed; the repair is not specific to the matching clinical dimension and tracks acoustic difficulty rather than prompt content (Appendix~\ref{app:additional}).

\subsection{Limitations}
\label{sec:limitations}

Our evaluation is restricted to the Speech Accessibility Project (SAP) dataset, which, despite being the largest dysarthric speech corpus, presents several structural constraints: the clinical annotations are coarse, single-rater judgments on an ordinal 1--7 scale rather than standardized instruments (e.g., FDA-2), precluding inter-rater reliability analysis. The rated subset is roughly 3\% of the dataset and artificially balanced toward more severe and diverse speech; it overrepresents Parkinson's disease (67\%), provides limited Down syndrome (3\%), and contains no stroke samples with complete profiles. The exclusive focus on US English-speaking adults prevents generalization to pediatric or cross-linguistic populations.

The population-level improvement from context is modest and not significant in aggregate, and the fine-tuned systems respond primarily to context presence (Section~\ref{sec:analysis}). The supervised experiments cover two architectures, and the speech encoder remains frozen during fine-tuning; retraining it may be important for atypical acoustics. The difficulty-aware gate is computed with an oracle difficulty signal, so it quantifies the benefit available from selective context use rather than a deployable system. Cerebral palsy is the only etiology for which the context-dependent system slightly degrades, possibly reflecting inter-speaker articulatory variability that clinical profiles do not capture, and speakers with severe impairment are underrepresented in the rated subset ($N = 21$ utterances).

The scarcity of dedicated audio-native reasoning models limits the generality of our chain-of-thought findings, since fundamental reasoning limits are hard to separate from model-specific artifacts such as reasoning-chain leakage. Finally, we evaluate isolated utterances with text-based context only, leaving audio-based in-context learning with same-speaker references for future work.

\section{Conclusion}
\label{sec:conclusion}
We presented a systematic evaluation of whether audio-language models can leverage clinical context for dysarthric speech recognition. Across ten models and nine prompt conditions, frozen models do not reliably use such context, and counterfactual controls show the degradation tracks the mere presence of non-instructional prompt text rather than its accuracy. The three failure modes we identify suggest a limitation of training exposure rather than architecture.

Context-dependent fine-tuning, evaluated on two model families, supports this interpretation. Voxtral-Small reaches a WER of 0.066 (a 52\% relative reduction over the frozen baseline) while preserving performance when context is unavailable, and the pattern replicates on Gemma-4-E2B. Counterfactual controls show, however, that with the partial exception of the smaller Gemma model, correct and wrong clinical content are statistically equivalent for the fine-tuned systems: the models learn that context is present, not what it says. The benefit of context grows with acoustic difficulty and concentrates on the hardest utterances, and applying context only there, through a difficulty-aware gate, lowers WER by about 10\% relative over the audio-only baseline on both architectures. Clinical context is thus most valuable exactly where accessibility needs are greatest. Future work should explore training objectives that encourage models to condition on individual clinical features, and no-reference difficulty estimators that make the gate deployable.

Any attempt to exploit clinical context must also weigh its ethical implications. Clinical labels may encourage a deficit-oriented framing of atypical speech, so such information should be optional, user-controlled support rather than a requirement for access; and because language models inherit social biases, identifying a speaker as disabled may itself affect transcription or downstream interpretation harmfully. 
A further risk is hallucinated correction: when a model reshapes speech toward a more fluent form, it may change the speaker’s intended meaning, which is especially problematic in AAC settings.

Two broader implications follow. Accessibility for dysarthric speech will not follow automatically from scaling general-purpose audio-language models \citep{hoffman2014nhis, jaddoh2025overcomingspeechbarriers}, even when contextual information is available, and this evaluation of open-weight models offers a reference point for future work, placing responsibility on developers of large-scale systems to evaluate accessibility directly.

\paragraph{Acknowledgments.}\label{sec:ack}
We obtained official approval from the authors of the Speech Accessibility Project to evaluate audio-language models that do not violate the data redistribution regulations, and gratefully acknowledge the University of Illinois Urbana-Champaign, Beckman Institute for Advanced Science and Technology, for their permission and support in this study.

\bibliography{colm2026_conference}
\bibliographystyle{colm2026_conference}
\newpage
\appendix

\section{Prompt Templates}
\label{app:prompts}
All prompt conditions share a fixed preamble and task instruction. Context blocks between them are composed modularly depending on the condition.

\subsection*{Prompt Structure}

\begin{tcolorbox}[
  colback=gray!5,
  colframe=gray!60,
  title=Prompt Structure,
  coltitle=black,
  fonttitle=\bfseries,
  boxrule=0.8pt,
  arc=4pt
]

\textbf{Preamble} \hfill {\small (constant across all conditions)}

\vspace{4pt}
\hrule
\vspace{4pt}

\textbf{Context Blocks} \hfill {\small (composable, toggleable)}
\begin{itemize}
  \item \texttt{condition} — diagnosis + guidance
  \item \texttt{pathology} — clinical speech ratings
  \item \texttt{follow\_up} — correction from prior pass
\end{itemize}

\vspace{4pt}
\hrule
\vspace{4pt}

\textbf{Task Instruction} \hfill {\small (constant across all conditions)}

\end{tcolorbox}

\noindent The \textbf{preamble} and \textbf{task instruction} are invariant.

\begin{quote}
\small
\textbf{Preamble:} \textit{You are transcribing speech from a person with a speech disorder. The audio may contain atypical pronunciation, rhythm, or voice quality. Use the provided clinical context to interpret ambiguous segments.}

\vspace{1mm}
\textbf{Task instruction:} \textit{Transcribe the audio in English. Output ONLY the transcription, no explanations.}
\end{quote}

\subsection*{P0, Zero-Shot Baseline}

No clinical context is provided. The model receives only the audio and a plain transcription instruction.

\begin{tcolorbox}[colback=gray!5, colframe=gray!50, coltitle=black, title=P0: Zero-Shot, fonttitle=\bfseries\small]
\small
Transcribe this audio in English. Output only the transcription, no explanations.
\end{tcolorbox}

\subsection*{P1, Condition Awareness}

The model is informed of the speaker's diagnosis and receives condition-specific guidance about expected speech characteristics. No per-sample clinical ratings are included.

\begin{tcolorbox}[colback=blue!3, colframe=blue!40,coltitle=black, title=P1: Condition Awareness (Cerebral Palsy example), fonttitle=\bfseries\small]
\small
You are transcribing speech from a person with a speech disorder. The audio may contain atypical pronunciation, rhythm, or voice quality. Use the provided clinical context to interpret ambiguous segments.

\vspace{2mm}
The speaker has Cerebral Palsy.\\
Cerebral Palsy often causes imprecise consonants, distorted vowels, and irregular speech rhythm. Words may sound slurred or have unusual stress patterns. Focus on the intended words rather than the surface-level distortions.

\vspace{2mm}
Transcribe the audio in English. Output ONLY the transcription, no explanations.
\end{tcolorbox}

\subsection*{P2, Full Clinical Profile}

In addition to the condition block from P1, the model receives per-sample clinical speech ratings across all annotation dimensions, sorted by severity.

\begin{tcolorbox}[colback=green!3, colframe=green!40,coltitle=black, title=P2: Full Clinical Profile (Cerebral Palsy example), fonttitle=\bfseries\small]
\small
You are transcribing speech from a person with a speech disorder. The audio may contain atypical pronunciation, rhythm, or voice quality. Use the provided clinical context to interpret ambiguous segments.

\vspace{2mm}
The speaker has Cerebral Palsy.\\
Cerebral Palsy often causes imprecise consonants, distorted vowels, and irregular speech rhythm. Words may sound slurred or have unusual stress patterns. Focus on the intended words rather than the surface-level distortions.

\vspace{2mm}
Clinical speech profile (scale 1--7, 1=normal, 7=most severe):\\
\hspace*{4mm}-- Naturalness: 5/7 (moderate-severe)\\
\hspace*{4mm}-- Imprecise consonants: 4/7 (moderate)\\
\hspace*{4mm}-- Monopitch: 4/7 (moderate)\\
\hspace*{4mm}-- Distorted vowels: 3/7 (mild-moderate)\\
\hspace*{4mm}-- Harsh voice: 3/7 (mild-moderate)\\
\hspace*{4mm}-- Slow rate: 3/7 (mild-moderate)\\
\hspace*{4mm}-- Intelligibility: 2/7 (mild)\\
\hspace*{4mm}-- Low pitch: 1/7 (normal)

\vspace{2mm}
Transcribe the audio in English. Output ONLY the transcription, no explanations.
\end{tcolorbox}

\subsection*{P3, Full Profile with Follow-Up Correction}

Extends P2 with additional context from a prior transcription attempt, enabling iterative self-correction.

\begin{tcolorbox}[colback=orange!3, colframe=orange!40,coltitle=black, title=P3: Full Profile + Follow-Up (ALS example), fonttitle=\bfseries\small]
\small
You are transcribing speech from a person with a speech disorder. The audio may contain atypical pronunciation, rhythm, or voice quality. Use the provided clinical context to interpret ambiguous segments.

\vspace{2mm}
The speaker has ALS.\\
ALS progressively weakens speech muscles, leading to slow, effortful speech with breathy or strained voice quality. Words may be prolonged or have nasal quality. Listen for the intended message through the motor speech difficulties.

\vspace{2mm}
Clinical speech profile (scale 1--7, 1=normal, 7=most severe):\\
\hspace*{4mm}-- Imprecise consonants: 5/7 (moderate-severe)\\
\hspace*{4mm}-- Breathiness: 4/7 (moderate)\\
\hspace*{4mm}-- Slow rate: 4/7 (moderate)\\
\hspace*{4mm}-- Hypernasality: 3/7 (mild-moderate)

\vspace{2mm}
Additional context from prior attempt: Previous transcription produced ``I quite assure you I am too kind''. The correct phrasing is ``I'm quite sure you are too kind-hearted''.

\vspace{2mm}
Transcribe the audio in English. Output ONLY the transcription, no explanations.
\end{tcolorbox}

\subsection*{Ablation Sub-Conditions}

The clinical ratings in P2 are organized into three tiers by ASR relevance. Sub-conditions isolate individual tiers to measure their contribution:

\begin{table}[h]
\centering
\small
\begin{tabular}{llp{7cm}}
\toprule
\textbf{ID} & \textbf{Ratings Included} & \textbf{Dimensions} \\
\midrule
P2a & Tier~1 only & Imprecise consonants, distorted vowels, repeated/prolonged phonemes, speech rate, stress patterns, nasality, short phrases, inappropriate silences \\
P2b & Tier~2 only & Harsh/strained/breathy voice, voice tremor, pitch breaks, monopitch, monoloudness, low pitch, prolonged intervals \\
P2c & Tier~1 + 2 & All actionable and voice quality dimensions \\
P2d & Condensed, all tiers & Compact key-value format: \texttt{condition: Cerebral Palsy; speech\_ratings: Imprecise consonants=4/7, Distorted vowels=3/7, ...} \\
P2e & Condensed, Tier~1 & Compact format, ASR-actionable dimensions only \\
\bottomrule
\end{tabular}
\caption{Ablation sub-conditions for the clinical ratings block. Tier~3 (meta-level judgments: naturalness, intelligibility, other) is included only in P2 and P2d.}
\label{tab:ablation_conditions}
\end{table}

\subsection*{Condition-Specific Guidance}

The condition block in P1--P3 includes a diagnosis-specific description tailored to each etiology in the SAP dataset:

\begin{table}[t]
\centering
\small
\begin{tabular}{lp{10cm}}
\toprule
\textbf{Etiology} & \textbf{Guidance Text} \\
\midrule
Cerebral Palsy & Cerebral Palsy often causes imprecise consonants, distorted vowels, and irregular speech rhythm. Words may sound slurred or have unusual stress patterns. Focus on the intended words rather than the surface-level distortions. \\
\addlinespace
ALS & ALS progressively weakens speech muscles, leading to slow, effortful speech with breathy or strained voice quality. Words may be prolonged or have nasal quality. Listen for the intended message through the motor speech difficulties. \\
\addlinespace
Parkinson's Disease & Parkinson's Disease typically causes reduced loudness, monotone pitch, and sometimes rapid or mumbled speech. Words may run together or trail off. Pay close attention to softly spoken or rushed segments. \\
\addlinespace
Down syndrome & Down syndrome can affect speech clarity through imprecise articulation and irregular speech rhythm. The speaker may have difficulty with certain consonant clusters. Focus on the overall message and common word patterns. \\
\addlinespace
Stroke & Stroke can cause various speech difficulties including slurred speech, word-finding pauses, or sound substitutions. The speaker's intended words may differ from how they sound on the surface. \\
\bottomrule
\end{tabular}
\caption{Etiology-specific guidance text included in the condition block (P1--P3).}
\label{tab:condition_guidance}
\end{table}

\subsection*{Counterfactual Control Conditions}

Three control conditions preserve the structure and length of the full clinical profile (P2) while manipulating its content (results in Section~\ref{sec:results_counterfactual} and Section~\ref{sec:results_supervised}).

\begin{itemize}[leftmargin=1.4em]
  \item \textbf{Wrong clinical profile.} The diagnosis and per-sample ratings are replaced with those of a speaker with a different etiology, via cyclic rotation across etiologies (e.g., a cerebral palsy speaker receives an ALS speaker's diagnosis, guidance, and ratings). Format, length, and rating-scale statistics are preserved; only the correctness of the content is destroyed.
  \item \textbf{Length-matched random text.} The condition and ratings blocks are replaced by fluent text with no relation to the task or the speaker, matched in length to the full clinical profile.
  \item \textbf{Length-matched instruction-only filler.} The condition and ratings blocks are replaced by generic transcription instructions (restating output-format requirements), matched in length and containing no non-instructional content.
\end{itemize}

\begin{table}[t]
\caption{Prompt condition hierarchy.  Each level builds on the previous
by adding clinical information.  All conditions share the same
$N=11{,}218$ evaluation samples, enabling paired comparisons.
Full example prompts can be found in Appendix~\ref{app:prompts}.}
\label{tab:conditions}
\centering
\small
 
\begin{tcolorbox}[
  enhanced,
  colback=white,
  colframe=outerframe,
  coltitle=white,
  fonttitle=\bfseries\small,
  title={Prompt Condition Hierarchy},
  boxrule=0.9pt,
  arc=4pt,
  top=6pt, bottom=6pt, left=26pt, right=6pt,
  width=\columnwidth,
  overlay={
    \draw[-stealth, line width=1.4pt, outerframe]
      ([xshift=19pt, yshift=-14pt] interior.north west)
      -- ([xshift=19pt, yshift=14pt] interior.south west);
    \node[rotate=90, anchor=south,
          font=\footnotesize\itshape,
          text=outerframe]
      at ([xshift=17pt]$(interior.north west)!0.5!(interior.south west)$)
      {increasing clinical information};
  }
]
 
\begin{tcolorbox}[
  enhanced,
  colback=p0gray!5,
  colframe=p0gray!50,
  coltitle=white,
  fonttitle=\bfseries\small,
  title={\textsf{P0}\enspace Zero-context control},
  boxrule=0.6pt, arc=3pt,
  top=2pt, bottom=2pt, left=4pt, right=4pt,
  fontupper=\small
]
Transcription instruction only; no clinical metadata.
Matched baseline for all paired comparisons.
\end{tcolorbox}
 
\vspace{4pt}
 
\begin{tcolorbox}[
  enhanced,
  colback=p1blue!3,
  colframe=p1blue!50,
  coltitle=white,
  fonttitle=\bfseries\small,
  title={\textsf{P1}\enspace Diagnosis guidance},
  boxrule=0.6pt, arc=3pt,
  top=2pt, bottom=2pt, left=4pt, right=4pt,
  fontupper=\small
]
\textcolor{p0gray}{\textsf{P0}}\,+ diagnosis label and
condition-specific guidance
(e.g.\ \textit{``The speaker has Cerebral Palsy.
CP often causes imprecise consonants\ldots''}).
No per-sample ratings.
\end{tcolorbox}
 
\vspace{4pt}
 
\begin{tcolorbox}[
  enhanced,
  colback=p2green!3,
  colframe=p2green!50,
  coltitle=white,
  fonttitle=\bfseries\small,
  title={\textsf{P2}\enspace Clinical profiles
         \normalfont\small\enspace (six ablation variants)},
  boxrule=0.6pt, arc=3pt,
  top=2pt, bottom=3pt, left=4pt, right=4pt,
  fontupper=\small
]
\textcolor{p1blue}{\textsf{P1}}\,+ per-sample clinician ratings
(scale 1--7), in six configurations:\par
\vspace{3pt}
{\footnotesize
\begin{tabular}{@{}l@{\ \ }p{0.76\columnwidth}@{}}
  \textbullet\ \textbf{Full profile}
    & Speech-production + voice-quality + overall meta-ratings. \\[0.5pt]
  \textbullet\ \textbf{Speech prod.}
    & Articulation, timing, stress, resonance, phrasing only. \\[0.5pt]
  \textbullet\ \textbf{Voice quality}
    & Phonation, pitch, loudness, tremor only. \\[0.5pt]
  \textbullet\ \textbf{Speech\,+\,voice}
    & Both of the above; no meta-ratings. \\[0.5pt]
  \textbullet\ \textbf{Condensed full}
    & Compact key-value format of all ratings. \\[0.5pt]
  \textbullet\ \textbf{Condensed speech}
    & Compact format, production ratings only. \\
\end{tabular}}
\end{tcolorbox}
 
\vspace{4pt}
 
\begin{tcolorbox}[
  enhanced,
  colback=p3orange!4,
  colframe=p3orange!50,
  coltitle=white,
  fonttitle=\bfseries\small,
  title={\textsf{P3}\enspace Follow-up correction},
  boxrule=0.6pt, arc=3pt,
  top=2pt, bottom=2pt, left=4pt, right=4pt,
  fontupper=\small
]
\textcolor{p2green}{\textsf{P2}}\,(full)\,+ a prior-pass transcript
for iterative correction.
Most information-rich prompting scenario.
\end{tcolorbox}
 
\end{tcolorbox}
 
\end{table}

\section{Metrics}
\label{app:metrics}
\subsection{Evaluation Metrics}
\label{app:standard_metrics}

Word error rate (WER) and character error rate (CER) are the primary metrics. WER is defined as
\[
    \mathrm{WER} = \frac{S + D + I}{N}
\]
where $S$, $D$, $I$ are substitutions, deletions, and insertions respectively, and $N$ is the number of words in the reference. CER is computed analogously at the character level.

\paragraph{Text normalization.} Following the SAP benchmark protocol~\citep{zheng2025interspeech2025speechaccessibility}, both hypotheses and references are normalized using Whisper's \texttt{EnglishTextNormalizer}~\citep{radford2022robustspeechrecognitionlargescale}. This includes number-to-word expansion, contraction normalization, British-to-American spelling conversion, filler word removal, and stripping of bracket-delimited prompt prefixes that appear in spontaneous speech references (e.g., \texttt{[Tell us about your hobbies.]}). Hypotheses are additionally truncated to 512 words after normalization.

\paragraph{Long-audio decoding.} Models with a fixed 30\,s audio input limit (Whisper and the Gemma-4 family) would otherwise silently truncate the 1{,}081 rated samples longer than 30\,s. For these models, long samples are decoded in segments of at most 30\,s, split at forced-alignment word gaps, with the same prompt applied to every segment and the segment transcripts concatenated.

\paragraph{Dual-reference minimum.} The SAP dataset provides two reference transcriptions per utterance. The verbatim \textit{transcript}~$r_t$ preserves disfluency markers (e.g., \texttt{(um)}, \texttt{(uh)}, incomplete words such as \texttt{(w-)}) and the \textit{clean transcript}~$r_c$ has these markers removed. Since models are not expected to reproduce disfluency annotations, per-sample WER is computed against both references and the minimum is taken.
\[
    \mathrm{WER}_{\mathrm{min}}^{(i)} = \min\!\big(\mathrm{WER}(\hat{y}_i,\, r_t^{(i)}),\; \mathrm{WER}(\hat{y}_i,\, r_c^{(i)})\big)
\]
CER\textsubscript{min} is defined analogously.

\paragraph{Clip-at-one.} Per-sample WER and CER are clipped at 1.0 before averaging.

\[\overline{\mathrm{WER}} = \frac{1}{N} \sum_{i=1}^{N} \min\!\big(\mathrm{WER}_{\mathrm{min}}^{(i)},\; 1.0\big)\]
This convention, adopted from the SAP challenge, prevents individual outliers from dominating the aggregate metric. Such outliers arise frequently in practice, as autoregressive models, particularly under sampling-based decoding or with long context prompts, can enter repetitive generation loops (e.g., producing ``no no no no\ldots'' hundreds of times), yielding per-sample $\mathrm{WER} \gg 1$. Since the severity of these hallucinations depends on the maximum generation length rather than on transcription quality, clipping ensures that a single degenerate sample does not overshadow systematic improvements across thousands of correctly transcribed utterances. We additionally report unclipped WER and hallucination rates (percentage of samples with $\mathrm{WER}_\mathrm{raw} > 1.0$) to characterize these failure modes separately.

\subsection{SemScore}
\label{app:semscore}

WER treats all word-level mismatches equally, penalizing semantically equivalent paraphrases (e.g., ``what is'' vs.\ ``what's'') as harshly as meaning-altering errors (e.g.\ ``what is'' vs.\ ``watt is''). To capture semantic similarity, we additionally compute SemScore~\citep{phukon2025aligningasrevaluationhuman}, a composite metric that achieved $\rho = 0.89$ correlation with human intelligibility judgments on dysarthric speech in the SAP evaluation. SemScore combines, via learned weights, three sub-metrics.
\[
    \mathrm{SemScore} = 0.40 \cdot \hat{s}_{\mathrm{NLI}} + 0.28 \cdot \hat{s}_{\mathrm{BERT}} + 0.32 \cdot \hat{s}_{\mathrm{phon}}
\]
where each $\hat{s}$ is min-max normalized to $[0, 1]$, consistently with the fixed thresholds from the SAP protocol.

\begin{itemize}
    \item \textbf{NLI entailment} ($s_{\mathrm{NLI}}$) measures bidirectional textual entailment probability from a RoBERTa-large model fine-tuned on SNLI, MNLI, FEVER, and ANLI~\citep{nie2020adversarialnlinewbenchmark}. The score averages $P(\mathrm{entailment} \mid r, \hat{y})$ and $P(\mathrm{entailment} \mid \hat{y}, r)$, capturing whether prediction and reference mutually imply each other regardless of exact wording.
    \item \textbf{BERTScore} ($s_{\mathrm{BERT}}$) measures token-level contextual embedding similarity (F1) using RoBERTa-large with baseline rescaling~\citep{zhang2020bertscoreevaluatingtextgeneration}.
    \item \textbf{Phonetic similarity} ($s_{\mathrm{phon}}$) computes Jaro-Winkler similarity between Soundex encodings of prediction and reference, capturing phonetic resemblance independent of orthographic variation.
\end{itemize}

\noindent Following the dual-reference convention, SemScore is computed against both $r_t$ and $r_c$, and the per-sample \textit{maximum} is taken (higher SemScore is better, opposite to WER).\\
\[\mathrm{SemScore}_{\mathrm{max}}^{(i)} = \max\!\big(\mathrm{SemScore}(\hat{y}_i,\, r_t^{(i)}),\; \mathrm{SemScore}(\hat{y}_i,\, r_c^{(i)})\big)\]
We report SemScore on a 0--100 scale. Unlike WER, SemScore is not subject to the hallucination inflation problem, as a repetitive output receives low semantic similarity scores regardless of length, making it a natural complement to the clipped WER metric.

\subsection{Statistical Testing}
\label{app:stats}
Paired frozen-model comparisons use Wilcoxon signed-rank tests with Benjamini--Hochberg FDR correction across the 50 model~$\times$~condition comparisons (Section~\ref{sec:results_conditioning}). For the correct-versus-wrong comparisons on the fine-tuned systems (Section~\ref{sec:results_supervised}) we test \emph{equivalence} directly with two one-sided tests (TOST) on paired per-utterance WER differences, using an equivalence bound of $\pm 0.005$ WER; $p < 0.05$ rejects the hypothesis that the true difference exceeds the bound. Speaker-clustered 95\% confidence intervals are obtained by bootstrap resampling at the speaker level, which accounts for the within-speaker correlation of per-utterance WER. The difficulty analysis (Section~\ref{sec:results_difficulty}) bins utterances by the audio-only LoRA system's WER and reports, within each bin, the paired difference between the audio-only system and the context-dependent system evaluated with the full clinical profile.

\begin{table}[t]
\section{Supporting Tables}
\label{app:main_tables}

\centering
\caption{Zero-shot baseline WER on the SAP rated subset ($N = 11{,}218$). ``Promptable'' indicates whether the model accepts text instructions alongside audio. $^{\ast}$Audio Flamingo~3's high WER is due to non-transcription output without task instruction; the basic transcription instruction reduces it to 0.162 (see Section~\ref{sec:results_conditioning}).}
\label{tab:baselines}
\small
\begin{tabular}{lcr}
\toprule
Model & Promptable & WER \\
\midrule
Qwen3-ASR-1.7B          & No  & 0.1338 \\
Qwen3-Omni-30B          & Yes & 0.1377 \\
Voxtral-Small-24B       & Yes & 0.1388 \\
Whisper-large-v3        & No  & 0.1443 \\
Cohere Transcribe       & No  & 0.1448 \\
Voxtral-Mini-3B         & Yes & 0.1531 \\
Phi-4 Multimodal        & Yes & 0.1577 \\
Parakeet-TDT-1.1B       & No  & 0.1585 \\
Qwen3-Thinking-30B      & Yes & 0.1650 \\
Granite-Speech 4.1-2B NAR & No  & 0.1667 \\
MiniCPM-o 4.5           & Yes & 0.1812 \\
MiniCPM-o Think         & Yes & 0.1841 \\
Granite-Speech 4.1-2B+  & Yes & 0.1909 \\
Qwen2-Audio-7B          & Yes & 0.2013 \\
Granite-Speech 3.3-8B   & Yes & 0.2366 \\
Audio-Reasoner          & Yes & 0.2400 \\
Gemma-4-4B              & Yes & 0.2976 \\
Gemma-4-4B Think        & Yes & 0.3018 \\
Gemma-4-2B              & Yes & 0.3022 \\
Ultravox-Nemo 12B       & Yes & 0.3123 \\
Wav2Vec2-Robust         & No  & 0.3264 \\
Gemma-4-12B Think       & Yes & 0.3881 \\
Wav2Vec2                & No  & 0.3930 \\
Gemma-4-12B             & Yes & 0.4122 \\
Audio Flamingo 3$^{\ast}$ & Yes & 0.4955 \\
\bottomrule
\end{tabular}
\end{table}

\begin{table*}[t]
\centering
\caption{WER across clinical prompt conditions ($N = 11{,}218$ matched samples per model). Columns are prompt conditions: P0 (zero-context control), P1 (diagnosis guidance), P2 (full clinical profile), P2a--P2e (profile ablations: speech-production, voice-quality, speech+voice, condensed-full, condensed-speech), and P3 (follow-up correction); see Table~\ref{tab:conditions}. Bold marks each model's lowest (best) WER. No model improves meaningfully with clinical context. $^{\ast}$Audio Flamingo~3's zero-context (P0) WER of 0.162 reflects a large drop from its unprompted WER of 0.495; this is a format-correction effect (see Section~\ref{sec:results_conditioning}).}
\label{tab:conditioning}
\small
\centering
\begin{tabular}{lccccccccc}
\toprule
Model & P0 & P1 & P2 & P2a & P2b & P2c & P2d & P2e & P3 \\
\midrule
AF3$^{\ast}$      & 0.1616 & \textbf{0.1615} & 0.1627 & 0.1624 & 0.1621 & 0.1626 & 0.1637 & 0.1635 & 0.1625 \\
Gemma-4-4B        & \textbf{0.2925} & 0.3245 & 0.3412 & 0.3337 & 0.3331 & 0.3405 & 0.3190 & 0.3167 & 0.3075 \\
Gemma-4-4B Think  & \textbf{0.3018} & 0.5095 & 0.5039 & 0.5108 & 0.5136 & 0.5147 & 0.5036 & 0.5076 & 0.4355 \\
Gemma-4-12B       & \textbf{0.4122} & 0.5298 & 0.5005 & 0.5282 & 0.5440 & 0.5203 & 0.4468 & 0.4665 & 0.4195 \\
Granite-3.3-8B    & 0.2366 & 0.2127 & 0.2133 & 0.2160 & 0.2174 & 0.2116 & 0.2059 & \textbf{0.2032} & 0.2156 \\
Granite-4.1-2B+   & \textbf{0.1909} & 0.2023 & 0.2058 & 0.2022 & 0.2027 & 0.2047 & 0.2022 & 0.2008 & 0.2168 \\
MiniCPM-o Think   & 0.1839 & \textbf{0.1779} & 0.1783 & 0.1790 & 0.1794 & 0.1789 & 0.1805 & 0.1800 & 0.1789 \\
Phi-4             & 0.1568 & 0.1566 & 0.1572 & 0.1550 & \textbf{0.1524} & 0.1557 & 0.1616 & 0.1626 & 0.1592 \\
Qwen2-Audio       & \textbf{0.2014} & 0.2320 & 0.2524 & 0.2578 & 0.2528 & 0.2556 & 0.2396 & 0.2479 & 0.2572 \\
Qwen3-Omni        & \textbf{0.1377} & 0.1409 & 0.1415 & 0.1412 & 0.1415 & 0.1414 & 0.1429 & 0.1426 & 0.1416 \\
Qwen3-Think       & \textbf{0.1650} & 0.1709 & 0.1722 & 0.1721 & 0.1747 & 0.1723 & 0.1860 & 0.1814 & 0.1662 \\
Voxtral-S         & \textbf{0.1390} & 0.1486 & 0.1891 & 0.1788 & 0.1819 & 0.1866 & 0.1530 & 0.1474 & 0.1726 \\
\bottomrule
\end{tabular}
\end{table*}

\begin{table*}[t]
\centering
\caption{Paired comparison from zero-context control to full clinical profile ($N = 11{,}218$). Positive delta indicates degradation. Bold indicates largest change. $p$-values use Benjamini--Hochberg false-discovery-rate (FDR) correction.}
\label{tab:paired_headline}
\small
\resizebox{\textwidth}{!}{
\begin{tabular}{lrrrccrr}
\toprule
Model & Ctrl WER & Profile WER & $\Delta$ & Cohen $d$ & Effect & $p$ (FDR) & Degraded \\
\midrule
Audio Flamingo 3    & 0.1616 & 0.1627 & +0.0011 & 0.010 & negligible & $<$0.0001 & 11.7\% \\
MiniCPM-o Think     & 0.1839 & 0.1783 & $-$0.0056 & $-$0.040 & negligible & 0.3865       & 12.8\% \\
Phi-4               & 0.1568 & 0.1572 & +0.0004 & 0.003 & negligible & $<$0.0001 & 13.1\% \\
Qwen3-Omni-30B      & 0.1377 & 0.1415 & +0.0038 & 0.035 & negligible & $<$0.0001 & 12.3\% \\
Voxtral-Small-24B   & 0.1390 & 0.1891 & +0.0501 & 0.223 & small      & $<$0.0001 & 15.8\% \\
Qwen2-Audio-7B      & 0.2014 & 0.2524 & +0.0510 & 0.242 & small      & $<$0.0001 & 25.3\% \\
Qwen3-Thinking-30B  & 0.1650 & 0.1722 & +0.0072 & 0.049 & negligible & $<$0.0001 & 10.7\% \\
Gemma-4-4B          & 0.2925 & 0.3412 & +0.0487 & 0.212 & small      & $<$0.0001 & 19.4\% \\
Gemma-4-4B Think    & 0.3018 & 0.5039 & \textbf{+0.2021} & \textbf{0.575} & medium     & $<$0.0001 & 42.7\% \\
Gemma-4-12B         & 0.4122 & 0.5005 & +0.0883 & 0.233 & small      & $<$0.0001 & 27.4\% \\
Gemma-4-12B Think   & 0.3881 & 0.5015 & +0.1134 & 0.322 & small      & $<$0.0001 & 31.6\% \\
Granite-Speech 3.3-8B  & 0.2366 & 0.2133 & $-$0.0233 & $-$0.097 & negligible & $<$0.0001 & 12.3\% \\
Granite-Speech 4.1-2B+ & 0.1909 & 0.2058 & +0.0149 & 0.118 & negligible & $<$0.0001 & 17.7\% \\
\bottomrule
\end{tabular}
}
\end{table*}

\begin{table}[t]
\centering
\caption{WER delta (zero-context to full clinical profile) by severity bin. Positive values indicate degradation. Bold indicates largest change. Italics indicates largest improvement.}
\label{tab:severity_strat}
\small
\begin{tabular}{lrrr}
\toprule
Model & Mild (1--2) & Moderate (3--4) & Severe (5--7) \\
\midrule
AF3           & +0.002 & $-$0.000 & +0.035 \\
Phi-4         & $-$\textit{0.007} & +0.014 & +0.029 \\
MiniCPM-o Think & $-$0.003 & $-$\textit{0.011} & +0.004 \\
Qwen2-Audio   & +0.059 & +0.037 & +0.014 \\
Qwen3-Omni    & +0.003 & +0.005 & +0.035 \\
Qwen3-Think   & +0.006 & +0.009 & +0.047 \\
Gemma-4-4B    & +0.068 & +0.012 & +0.014 \\
Gemma-4-4B Think & \textbf{+0.224} & \textbf{+0.162} & \textbf{+0.093} \\
Voxtral-S     & +0.066 & +0.020 & +0.029 \\
Gemma-4-12B   & +0.107 & +0.054 & +0.081 \\
\bottomrule
\end{tabular}
\end{table}

\begin{table}[t]
\centering
\caption{WER delta (zero-context to full clinical profile) by etiology. Positive values indicate degradation. Bold indicates largest change. Italics indicates largest improvement.}
\label{tab:etiology_strat}
\small
\begin{tabular}{lrrrr}
\toprule
Model & ALS & CP & Down S. & Parkinson's \\
\midrule
AF3           & +0.004 & $-$0.003 & $-$0.009 & +0.002 \\
Phi-4         & +0.004 & $-$0.007 & $-$0.002 & +0.002 \\
MiniCPM-o Think & $-$\textit{0.013} & $-$0.018 & $-$\textit{0.032} & $-$0.000 \\
Qwen2-Audio   & +0.029 & +0.036 & +0.036 & +0.060 \\
Qwen3-Omni    & +0.001 & $-$\textit{0.019} & $-$0.009 & +0.010 \\
Qwen3-Think   & +0.001 & $-$0.010 & +0.031 & +0.011 \\
Gemma-4-4B    & +0.068 & +0.047 & +0.040 & +0.046 \\
Gemma-4-4B Think & \textbf{+0.175} & \textbf{+0.147} & \textbf{+0.196} & \textbf{+0.221} \\
Voxtral-S     & +0.057 & +0.046 & +0.077 & +0.048 \\
Gemma-4-12B   & +0.057 & +0.072 & +0.091 & +0.099 \\
\bottomrule
\end{tabular}
\end{table}

\begin{table}[t]
\centering
\caption{WER delta (zero-context to full clinical profile) by utterance category. Positive values indicate degradation. Bold indicates largest change. Italics indicates largest improvement.}
\label{tab:category_strat}
\small
\begin{tabular}{lrrrr}
\toprule
Model & Asst.\ Cmd & Non-spont. & Novel Sent. & Spont. \\
\midrule
AF3           & $-$0.002 & $-$0.005 & +0.008 & $-$0.002 \\
Phi-4         & $-$0.018 & $-$\textit{0.097} & +0.011 & +0.015 \\
MiniCPM-o Think & $-$\textit{0.019} & $-$0.011 & +0.007 & $-$\textit{0.004} \\
Qwen2-Audio   & +0.039 & +0.011 & +0.089 & +0.028 \\
Qwen3-Omni    & $-$0.007 & $-$0.054 & +0.020 & +0.002 \\
Qwen3-Think   & +0.001 & $-$0.032 & +0.011 & +0.012 \\
Gemma-4-4B    & +0.137 & +0.114 & +0.000 & $-$0.003 \\
Gemma-4-4B Think & \textbf{+0.256} & +0.119 & \textbf{+0.135} & \textbf{+0.215} \\
Voxtral-S     & +0.109 & \textbf{+0.124} & +0.008 & +0.024 \\
Gemma-4-12B   & +0.168 & +0.005 & +0.053 & +0.042 \\
\bottomrule
\end{tabular}
\end{table}

\begin{table}[t]
\centering
\caption{Change in error rates from zero-context control to full clinical profile ($\Delta$ rates normalised by total reference words). Bold indicates largest change. Italics indicates largest improvement.}
\label{tab:error_decomp}
\small
\begin{tabular}{lrrrr}
\toprule
Model & $\Delta$Sub & $\Delta$Ins & $\Delta$Del & $\Delta$Hit \\
\midrule
AF3         & $-$0.006 & +0.001 & +0.007 & $-$0.000 \\
Phi-4       & +0.025 & +0.043 & $-$0.007 & $-$0.018 \\
MiniCPM-o Think & +0.012 & +0.269 & $-$0.005 & $-$0.007 \\
Qwen2-Audio & +0.023 & +0.005 & +0.027 & $-$0.050 \\
Qwen3-Omni  & $-$0.008 & +0.030 & $-$0.002 & \textit{+0.009} \\
Qwen3-Think & +0.006 & $-$\textit{0.031} & +0.015 & $-$0.021 \\
Gemma-4-4B  & +0.008 & +0.063 & +0.000 & $-$0.008 \\
Gemma-4-4B Think & $-$\textit{0.010} & $-$0.025 & \textbf{+0.174} & $-$\textbf{0.165} \\
Voxtral-S   & +0.011 & +0.212 & $-$0.001 & $-$0.010 \\
Gemma-4-12B & \textbf{+0.031} & \textbf{+0.621} & $-$\textit{0.012} & $-$0.019 \\
\bottomrule
\end{tabular}
\end{table}

\section{Fine-Tuning Details}
\label{app:finetuning}

\subsection{Context-dependent Fine-Tuning Protocol}
We fine-tune Voxtral-Small-24B using LoRA (rank 16, $\alpha = 32$, dropout 0.05) applied to the query and value projection matrices of the language model layers. The audio encoder is frozen throughout.

Training uses 5-fold speaker-disjoint cross-validation. In each fold, approximately 80\% of speakers are used for training and the remaining 20\% (87--88 speakers per fold, 437 unique speakers total) are held out for evaluation. Folds are constructed so that no speaker appears in both training and evaluation within the same fold, preventing speaker-level data leakage.

The \textbf{audio-only} system trains exclusively on audio-transcription pairs with no clinical prompt. The \textbf{context-dependent} system trains on a mixture of three prompt formats per utterance, including audio-only (no clinical context), speech~+~voice profile (P2c), and condensed full profile (P2d). During each training epoch, the prompt format is sampled uniformly at random for each utterance, so the model sees the same audio with different context configurations across epochs. This exposure diversity is intended to teach the model to use context when present without degrading performance when it is absent.

Both systems are trained for 3 epochs per fold with a learning rate of $2 \times 10^{-4}$ (cosine schedule, 100 warmup steps), effective batch size of 8 (gradient accumulation over 4 steps), and bf16 mixed precision. We report the last checkpoint per fold. All evaluation results in the main text are pooled out-of-fold predictions across all five folds ($N = 11{,}218$ utterances).

\subsection{Gemma-4-E2B Replication}
\label{app:finetuning_gemma}
We replicate the protocol on Gemma-4-E2B (\texttt{google/gemma-4-E2B-it}; listed as Gemma-4-2B in Table~\ref{tab:models}) with the same LoRA configuration (rank 16, $\alpha = 32$, dropout 0.05) applied to the attention projections of the language-model layers ($\approx$3.1M trainable parameters, 0.06\% of the model), keeping the audio encoder frozen. Training uses the same 5-fold speaker-disjoint cross-validation splits and the same prompt-format mixture as the Voxtral-Small context-dependent system, with bf16 precision. Frozen Gemma-4-E2B reaches WER 0.302 at zero-context and degrades to 0.372 under the full clinical profile; after fine-tuning, both the audio-only and the context-dependent system reach WER 0.204, removing the degradation entirely (Section~\ref{sec:results_supervised}). The frozen evaluations use chunked decoding for long audio (Appendix~\ref{app:standard_metrics}); the LoRA evaluations predate this fix, and since chunking only lowers WER on the affected long samples, the reported fine-tuning improvements are, if anything, understated.

\subsection{Hardware}
Inference and fine-tuning for models up to 12B parameters were conducted on a single NVIDIA GeForce RTX 5090 (32\,GB VRAM) with an Intel Core i7-6900K. Larger models (Qwen3-Omni-30B, Qwen3-Thinking-30B) required a Project DIGITS developer kit with an NVIDIA GB10 (128\,GB unified memory).

\section{Counterfactual Control Results}
\label{app:counterfactual_tables}
Table~\ref{tab:frozen_controls} lists the frozen counterfactual control results (Section~\ref{sec:results_counterfactual}) and Table~\ref{tab:ft_counterfactual} the correct-versus-wrong comparison for the fine-tuned context-dependent systems (Section~\ref{sec:results_supervised}).

\begin{table}[t]
\centering
\caption{Frozen counterfactual controls (WER, $N = 11{,}218$ rated samples). Wrong clinical content is no better than correct content, length-matched random text degrades almost as much, and length-matched instruction-only filler sits on the audio-only baseline.}
\label{tab:frozen_controls}
\small
\begin{tabular}{lrr}
\toprule
Prompt condition & Voxtral-Small & Gemma-4-E2B \\
\midrule
Audio only (zero-context)            & 0.139 & 0.302 \\
Correct clinical profile             & 0.189 & 0.372 \\
Wrong clinical profile               & 0.190 & 0.383 \\
Random text (length-matched)         & 0.170 & 0.361 \\
Instruction filler (length-matched)  & 0.153 & 0.302 \\
\bottomrule
\end{tabular}
\end{table}

\begin{table}[t]
\centering
\caption{Correct vs.\ wrong clinical content for the fine-tuned context-dependent systems: WER difference (wrong $-$ correct, positive = correct better) with speaker-clustered 95\% CIs. Voxtral-Small is TOST-equivalent ($\pm 0.005$ bound, $p < 0.001$) at every condition. \textbf{Bold}: significant correct-over-wrong gain (Wilcoxon $p \le 0.005$).}
\label{tab:ft_counterfactual}
\small
\begin{tabular}{lcc}
\toprule
Eval condition & Voxtral-Small & Gemma-4-E2B \\
\midrule
Diagnosis guidance          & $-0.0004$ [$-0.0013$, $+0.0006$] & $+0.0011$ [$-0.0015$, $+0.0038$] \\
Full clinical profile       & $+0.0003$ [$-0.0007$, $+0.0012$] & $\mathbf{+0.0046}$ [$+0.0017$, $+0.0077$] \\
Speech production profile   & $+0.0001$ [$-0.0009$, $+0.0011$] & $\mathbf{+0.0034}$ [$+0.0007$, $+0.0062$] \\
Voice quality profile       & $-0.0007$ [$-0.0019$, $+0.0003$] & $+0.0030$ [$+0.0004$, $+0.0056$] \\
Speech + voice profile      & $+0.0003$ [$-0.0007$, $+0.0013$] & $+0.0015$ [$-0.0013$, $+0.0044$] \\
Condensed full profile      & $-0.0007$ [$-0.0018$, $+0.0004$] & $+0.0007$ [$-0.0019$, $+0.0032$] \\
Condensed speech production & $+0.0001$ [$-0.0009$, $+0.0010$] & $+0.0019$ [$-0.0006$, $+0.0043$] \\
\bottomrule
\end{tabular}
\end{table}

\section{Positioning Against Specialized SAP Systems}
\label{app:positioning}
\begin{figure}[t]
\centering
\includegraphics[width=0.62\columnwidth]{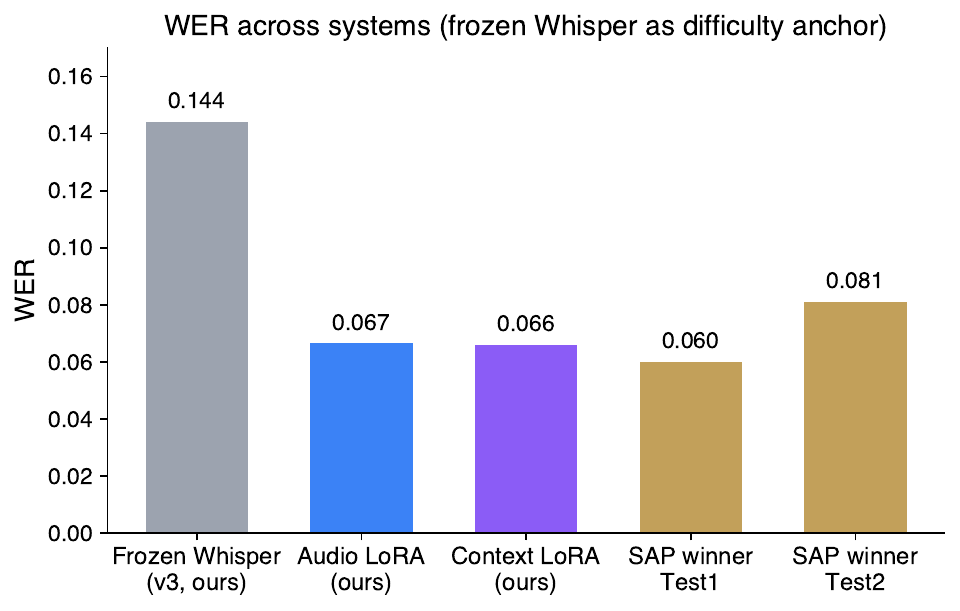}
\caption{WER across systems with frozen Whisper as a difficulty anchor. The systems use different SAP splits (our rated subset vs.\ challenge Test1/Test2) and normalization conventions; this is approximate positioning rather than a matched comparison. Challenge results from \citet{takahashi2025parakeet}.}
\label{fig:positioning}
\end{figure}

Figure~\ref{fig:positioning} places our fine-tuned systems next to the strongest published SAP system. The comparison is approximate: the SAP Challenge winner \citep{takahashi2025parakeet} is evaluated on the challenge private Test1/Test2 splits under full fine-tuning of Parakeet-TDT, whereas our systems are evaluated on the rated subset under parameter-efficient LoRA. The Gemini~2.5~Flash personalization system \citep{agarwal2025personalizer} reaches about 5\% WER with few-shot same-speaker enrollment but is omitted from the figure because its evaluation split is not directly comparable. A frozen Whisper anchor of comparable difficulty across splits (14--18\% WER) indicates that the evaluation sets are of similar hardness, so the comparison is approximate positioning rather than a matched benchmark.

\section{Full Paired Comparisons}
\label{app:full_paired}
Table~\ref{tab:full_paired} presents paired comparisons for the five primary prompt conditions across all models from the clinical context conditioning evaluation.
\begin{table*}[t]
\centering
\caption{Paired comparisons vs.\ zero-context control for the five primary prompt conditions. $p$-values use Benjamini--Hochberg false-discovery-rate (FDR) correction across all 50 comparisons. $N = 11{,}218$. Bold indicates the largest change ($|\Delta|$, $|d|$, and degradation rate) within each condition block.}
\label{tab:full_paired}
\small
\begin{tabular}{llrrcrr}
\toprule
Condition & Model & $\Delta$ & $d$ & Effect & $p$ (FDR) & Degraded \\
\midrule
\multirow{10}{*}{Diagnosis guidance}
  & AF3          & $-$0.0001 & $-$0.001 & negl. & $<$0.0001 & 10.9\% \\
  & Phi-4        & $-$0.0002 & $-$0.001 & negl. & $<$0.0001 & 13.0\% \\
  & MiniCPM-o Think & $-$0.0060 & $-$0.043 & negl. & 0.2112    & 12.3\% \\
  & Qwen2-Audio  & +0.0305   & +0.167   & negl. & $<$0.0001 & 22.7\% \\
  & Qwen3-Omni   & +0.0033   & +0.029   & negl. & $<$0.0001 & 12.1\% \\
  & Qwen3-Think  & +0.0059   & +0.039   & negl. & $<$0.0001 & 10.5\% \\
  & Gemma-4-4B   & +0.0321   & +0.165   & negl. & $<$0.0001 & 17.1\% \\
  & Gemma-4-4B Think & \textbf{+0.2076} & \textbf{+0.574} & medium & $<$0.0001 & \textbf{42.2\%} \\
  & Voxtral-S    & +0.0096   & +0.075   & negl. & $<$0.0001 & 11.1\% \\
  & Gemma-4-12B  & +0.1177   & +0.304   & small & $<$0.0001 & 29.0\% \\
\midrule
\multirow{10}{*}{Speech prod.\ profile}
  & AF3          & +0.0008   & +0.007   & negl. & $<$0.0001 & 11.4\% \\
  & Phi-4        & $-$0.0018 & $-$0.011 & negl. & $<$0.0001 & 13.4\% \\
  & MiniCPM-o Think & $-$0.0048 & $-$0.035 & negl. & 0.9011    & 13.2\% \\
  & Qwen2-Audio  & +0.0564   & +0.258   & small & $<$0.0001 & 26.6\% \\
  & Qwen3-Omni   & +0.0035   & +0.032   & negl. & $<$0.0001 & 12.0\% \\
  & Qwen3-Think  & +0.0071   & +0.047   & negl. & $<$0.0001 & 10.9\% \\
  & Gemma-4-4B   & +0.0413   & +0.192   & negl. & $<$0.0001 & 18.7\% \\
  & Gemma-4-4B Think & \textbf{+0.2090} & \textbf{+0.584} & medium & $<$0.0001 & \textbf{43.0\%} \\
  & Voxtral-S    & +0.0398   & +0.193   & negl. & $<$0.0001 & 14.8\% \\
  & Gemma-4-12B  & +0.1160   & +0.299   & small & $<$0.0001 & 29.8\% \\
\midrule
\multirow{10}{*}{Voice quality profile}
  & AF3          & +0.0004   & +0.004   & negl. & $<$0.0001 & 11.5\% \\
  & Phi-4        & $-$0.0044 & $-$0.028 & negl. & 0.0014    & 12.5\% \\
  & MiniCPM-o Think & $-$0.0045 & $-$0.032 & negl. & 0.7944    & 13.2\% \\
  & Qwen2-Audio  & +0.0514   & +0.251   & small & $<$0.0001 & 26.2\% \\
  & Qwen3-Omni   & +0.0038   & +0.035   & negl. & $<$0.0001 & 12.1\% \\
  & Qwen3-Think  & +0.0098   & +0.064   & negl. & $<$0.0001 & 10.9\% \\
  & Gemma-4-4B   & +0.0407   & +0.191   & negl. & $<$0.0001 & 18.3\% \\
  & Gemma-4-4B Think & \textbf{+0.2118} & \textbf{+0.586} & medium & $<$0.0001 & \textbf{43.6\%} \\
  & Voxtral-S    & +0.0429   & +0.203   & small & $<$0.0001 & 15.2\% \\
  & Gemma-4-12B  & +0.1319   & +0.335   & small & $<$0.0001 & 30.7\% \\
\midrule
\multirow{10}{*}{Condensed full profile}
  & AF3          & +0.0021   & +0.019   & negl. & $<$0.0001 & 11.2\% \\
  & Phi-4        & +0.0048   & +0.026   & negl. & $<$0.0001 & 11.8\% \\
  & MiniCPM-o Think & $-$0.0034 & $-$0.024 & negl. & 0.3184    & 13.0\% \\
  & Qwen2-Audio  & +0.0381   & +0.195   & negl. & $<$0.0001 & 23.6\% \\
  & Qwen3-Omni   & +0.0052   & +0.047   & negl. & $<$0.0001 & 12.9\% \\
  & Qwen3-Think  & +0.0210   & +0.118   & negl. & $<$0.0001 & 13.6\% \\
  & Gemma-4-4B   & +0.0265   & +0.145   & negl. & $<$0.0001 & 16.9\% \\
  & Gemma-4-4B Think & \textbf{+0.2018} & \textbf{+0.557} & medium & $<$0.0001 & \textbf{42.0\%} \\
  & Voxtral-S    & +0.0140   & +0.101   & negl. & $<$0.0001 & 11.8\% \\
  & Gemma-4-12B  & +0.0347   & +0.098   & negl. & $<$0.0001 & 21.7\% \\
\midrule
\multirow{10}{*}{Follow-up correction}
  & AF3          & +0.0009   & +0.011   & negl. & $<$0.0001 & 7.8\%  \\
  & Phi-4        & +0.0024   & +0.017   & negl. & $<$0.0001 & 9.2\%  \\
  & MiniCPM-o Think & $-$0.0050 & $-$0.035 & negl. & 0.7873    & 13.1\% \\
  & Qwen2-Audio  & +0.0557   & +0.300   & small & $<$0.0001 & 23.5\% \\
  & Qwen3-Omni   & +0.0039   & +0.045   & negl. & $<$0.0001 & 9.1\%  \\
  & Qwen3-Think  & +0.0012   & +0.013   & negl. & 0.3957    & 2.3\%  \\
  & Gemma-4-4B   & +0.0151   & +0.168   & negl. & $<$0.0001 & 8.0\%  \\
  & Gemma-4-4B Think & \textbf{+0.1337} & \textbf{+0.454} & small  & $<$0.0001 & \textbf{26.7\%} \\
  & Voxtral-S    & +0.0336   & +0.185   & negl. & $<$0.0001 & 11.0\% \\
  & Gemma-4-12B  & +0.0073   & +0.021   & negl. & 0.6453    & 11.8\% \\
\bottomrule
\end{tabular}
\end{table*}
\section{Context-dependent Evaluation Grid}
\label{app:ctx_cond_grid}
Table~\ref{tab:ctx_cond_grid} presents the full evaluation grid for the Voxtral-Small context-dependent system (5-fold CV), comparing against the audio-only system. Green triangles mark conditions where the context-dependent system beats the audio-only baseline.

\begin{table*}[t]
\centering
\caption{Context-dependent evaluation grid (Voxtral-Small, 5-fold CV, 11{,}218 pooled utterances). Audio-only WER is 0.0665 for reference. Bold indicates the best (lowest WER / largest improvement) value in each numeric column.}
\label{tab:ctx_cond_grid}
\small
\resizebox{\textwidth}{!}{
\begin{tabular}{lrrrr}
\toprule
Eval condition & Ctx-cond.\ WER & $\Delta$ vs zero-context & $\Delta$ vs audio-only & Rel.\% vs audio-only \\
\midrule
Zero-context control        & 0.0673 & ---       & +0.0008 & $-$1.2\% \\
Diagnosis guidance          & 0.0663 & $-$0.0010 & $-$0.0002 & +0.2\%  \\
Full clinical profile       & \textbf{0.0659} & \textbf{$-$0.0014} & \textbf{$-$0.0005} & \textbf{+0.8\%}  \\
Speech production profile   & 0.0664 & $-$0.0009 & $-$0.0000 & +0.0\%  \\
Voice quality profile       & 0.0662 & $-$0.0010 & $-$0.0002 & +0.3\%  \\
Speech + voice profile      & 0.0662 & $-$0.0010 & $-$0.0002 & +0.3\%  \\
Condensed full profile      & 0.0663 & $-$0.0010 & $-$0.0002 & +0.3\%  \\
Condensed speech prod.      & 0.0661 & $-$0.0012 & $-$0.0003 & +0.5\%  \\
\bottomrule
\end{tabular}
}

\end{table*}

\begin{table}[t]
\centering
\caption{Supervised fine-tuning results (5-fold CV, 437 speakers, 11{,}218 pooled utterances).}
\label{tab:supervised_summary}
\small
\begin{tabular}{llrr}
\toprule
System & Eval condition & WER & CER \\
\midrule
Voxtral-Small (frozen)         & Unprompted               & 0.1388 & 0.0980 \\
audio-only LoRA                & audio-only               & 0.0665 & 0.0506 \\
Context-cond.\ LoRA            & Zero-context control     & 0.0673 & 0.0511 \\
\textbf{Context-cond.\ LoRA}   & \textbf{Full clinical profile} & \textbf{0.0659} & \textbf{0.0497} \\
\bottomrule
\end{tabular}
\end{table}

\begin{table*}[t]
\centering
\caption{Headline paired comparisons for supervised systems (5-fold CV, 437 speakers). 95\% CI at the speaker level.}
\label{tab:supervised_paired}
\small
\resizebox{\textwidth}{!}{
\begin{tabular}{lrrrrrrr}
\toprule
Comparison & Base WER & New WER & $\Delta$ & Rel.\% & $p_{\text{spkr}}$ & 95\% CI (spkr) \\
\midrule
Frozen $\to$ audio-only LoRA                         & 0.1388 & 0.0665 & $-$0.0723 & +52.1 & $<$.0001 & ---           \\
audio-only $\to$ Context-cond.\ (full clin.\ prof.)  & 0.0665 & 0.0659 & $-$0.0005 & +0.8  & 0.5517   & [$-$.0015, +.0023] \\
Context-cond.\ zero-context $\to$ full clin.\ prof.\ (same weights)  & 0.0673 & 0.0659 & $-$0.0014 & +2.0  & 0.3149   & [$-$.0027, +.0007] \\
\bottomrule
\end{tabular}
}
\end{table*}

\begin{table}[t]
\centering
\caption{audio-only $\to$ context-dependent (full clinical profile) by severity bin.}
\label{tab:sub_severity}
\small
\begin{tabular}{lrrrrc}
\toprule
Severity & $N$ & audio-only & Ctx-cond. & $\Delta$ & $p_{\text{spkr}}$ \\
\midrule
Mild (1--2)     & 7{,}304 & 0.031 & 0.030 & $-$0.001 & \textbf{0.033} \\
Moderate (3--4) & 3{,}893 & 0.131 & 0.131 & +0.000   & 0.900 \\
Severe (5--7)   & 21      & 0.616 & 0.605 & $-$0.011 & 0.750 \\
\bottomrule
\end{tabular}
\end{table}

\begin{table}[t]
\centering
\caption{audio-only $\to$ context-dependent (full clinical profile) by etiology. $p$ values from Wilcoxon signed-rank test at the speaker level.}
\label{tab:sub_etiology}
\small
\begin{tabular}{lrrrrrc}
\toprule
Etiology & $N$ & Spkrs & audio-only & Ctx-cond. & $\Delta$ & $p_{\text{spkr}}$ \\
\midrule
Down syndrome       & 358   & 21  & 0.152 & 0.141 & $-$0.011 & \textbf{0.044} \\
ALS                 & 1{,}594 & 70  & 0.059 & 0.057 & $-$0.003 & 0.066 \\
Parkinson's disease & 7{,}504 & 249 & 0.037 & 0.036 & $-$0.001 & 0.231 \\
Cerebral palsy      & 1{,}762 & 97  & 0.180 & 0.185 & +0.006   & 0.121 \\
\bottomrule
\end{tabular}
\end{table}

\section{Models}

\begin{table}[ht]
\centering
\caption{Models evaluated for dysarthric speech transcription on the SAP dataset. Voxtral-Small-24B and Gemma-4-2B (Gemma-4-E2B) are additionally fine-tuned with LoRA (Section~\ref{sec:results_supervised}).}
\label{tab:models}
\resizebox{\textwidth}{!}{%
\begin{tabular}{llrl}
\toprule
\textbf{Model} & \textbf{Type} & \textbf{Params} & \textbf{HuggingFace Model ID} \\
\midrule
\multicolumn{4}{l}{\textit{Pure ASR}} \\
\midrule
Qwen3-ASR-1.7B          & Pure ASR  & 1.7B  & \texttt{Qwen/Qwen3-ASR-1.7B} \\
Granite-Speech 4.1-2B NAR & Pure ASR & 2B    & \texttt{ibm-granite/granite-speech-4.1-2b-nar} \\
Cohere Transcribe        & Pure ASR  & 2.0B  & \texttt{CohereLabs/cohere-transcribe-03-2026} \\
Whisper Large v3         & Pure ASR  & 1.5B  & \texttt{openai/whisper-large-v3} \\
NVIDIA Parakeet TDT 1.1B & Pure ASR  & 1.1B  & \texttt{nvidia/parakeet-tdt-1.1b} \\
wav2vec2-large-robust    & Pure ASR  & 0.3B  & \texttt{facebook/wav2vec2-large-robust-ft-libri-960h} \\
wav2vec2-large-960h      & Pure ASR  & 0.3B  & \texttt{facebook/wav2vec2-large-960h} \\
\midrule
\multicolumn{4}{l}{\textit{Instruct (audio + text input)}} \\
\midrule
Voxtral-Small-24B        & Instruct  & 24B   & \texttt{mistralai/Voxtral-Small-24B-2507} \\
Voxtral-Mini-3B          & Instruct  & 3B    & \texttt{mistralai/Voxtral-Mini-3B-2507} \\
Phi-4-multimodal         & Instruct  & 5.6B  & \texttt{microsoft/Phi-4-multimodal-instruct} \\
Qwen2-Audio-7B           & Instruct  & 7B    & \texttt{Qwen/Qwen2-Audio-7B-Instruct} \\
MiniCPM-o 4.5            & Instruct  & 9B    & \texttt{openbmb/MiniCPM-o-4\_5} \\
Qwen3-Omni-30B-A3B      & Instruct  & 30B\textsuperscript{$\dagger$} & \texttt{Qwen/Qwen3-Omni-30B-A3B-Instruct} \\
Ultravox-v0.4.1-Nemo     & Instruct  & 12B   & \texttt{fixie-ai/ultravox-v0\_4\_1-mistral-nemo} \\
NVIDIA Audio Flamingo 3  & Instruct  & 8B    & \texttt{nvidia/audio-flamingo-3-hf} \\
Granite-Speech 3.3-8B    & Instruct  & 8B    & \texttt{ibm-granite/granite-speech-3.3-8b} \\
Granite-Speech 4.1-2B+   & Instruct  & 2B    & \texttt{ibm-granite/granite-speech-4.1-2b-plus} \\
Gemma-4-2B               & Instruct  & 2B    & \texttt{google/gemma-4-E2B-it} \\
Gemma-4-4B               & Instruct  & 4B    & \texttt{google/gemma-4-E4B-it} \\
Gemma-4-12B\textsuperscript{$\ddagger$} & Instruct  & 12B   & \texttt{google/gemma-4-12B-it} \\
\midrule
\multicolumn{4}{l}{\textit{Chain-of-Thought (CoT)}} \\
\midrule
Qwen3-Omni-30B-A3B-Thinking & CoT  & 30B\textsuperscript{$\dagger$} & \texttt{Qwen/Qwen3-Omni-30B-A3B-Thinking} \\
MiniCPM-o 4.5 Think      & CoT   & 9B    & \texttt{openbmb/MiniCPM-o-4\_5} \\
Gemma-4-4B Think          & CoT   & 4B    & \texttt{google/gemma-4-E4B-it} \\
Gemma-4-12B Think\textsuperscript{$\ddagger$} & CoT   & 12B   & \texttt{google/gemma-4-12B-it} \\
Audio-Reasoner            & CoT   & 7B    & \texttt{zhifeixie/Audio-Reasoner} \\
\bottomrule
\end{tabular}%
}
\vspace{2mm}
\footnotesize{$\dagger$\,Mixture-of-Experts architecture with 3B active parameters per forward pass. $\ddagger$\,Gemma-4-12B uses a unified (encoder-free) multimodal architecture: audio is projected directly into the language model without a separate audio encoder, unlike the E2B/E4B variants.}
\end{table}

\paragraph{Prompt placement and the unified Gemma-4-12B.} Gemma-4-12B is the only \emph{unified} (encoder-free) model in our set: audio is projected directly into the language model without a separate audio encoder. Like the other promptable models it degrades under clinical prompting supplied in the user turn (Table~\ref{tab:conditioning}), but the degradation is unusually large. In exploratory runs we observed only minor differences between placing the clinical prompt in the user turn versus a system turn for most models, whereas for the unified 12B a system-turn placement substantially reduced the degradation. We did not pursue this further; the effect may relate to how the unified architecture routes system versus user tokens, which we leave to future mechanistic analysis. Throughout the paper we report the user-turn condition, as it most closely reflects how clinical context would be supplied at deployment.

\section{Model Size and Performance}
\label{app:size}
\begin{figure}[t]
\centering
\includegraphics[width=\columnwidth]{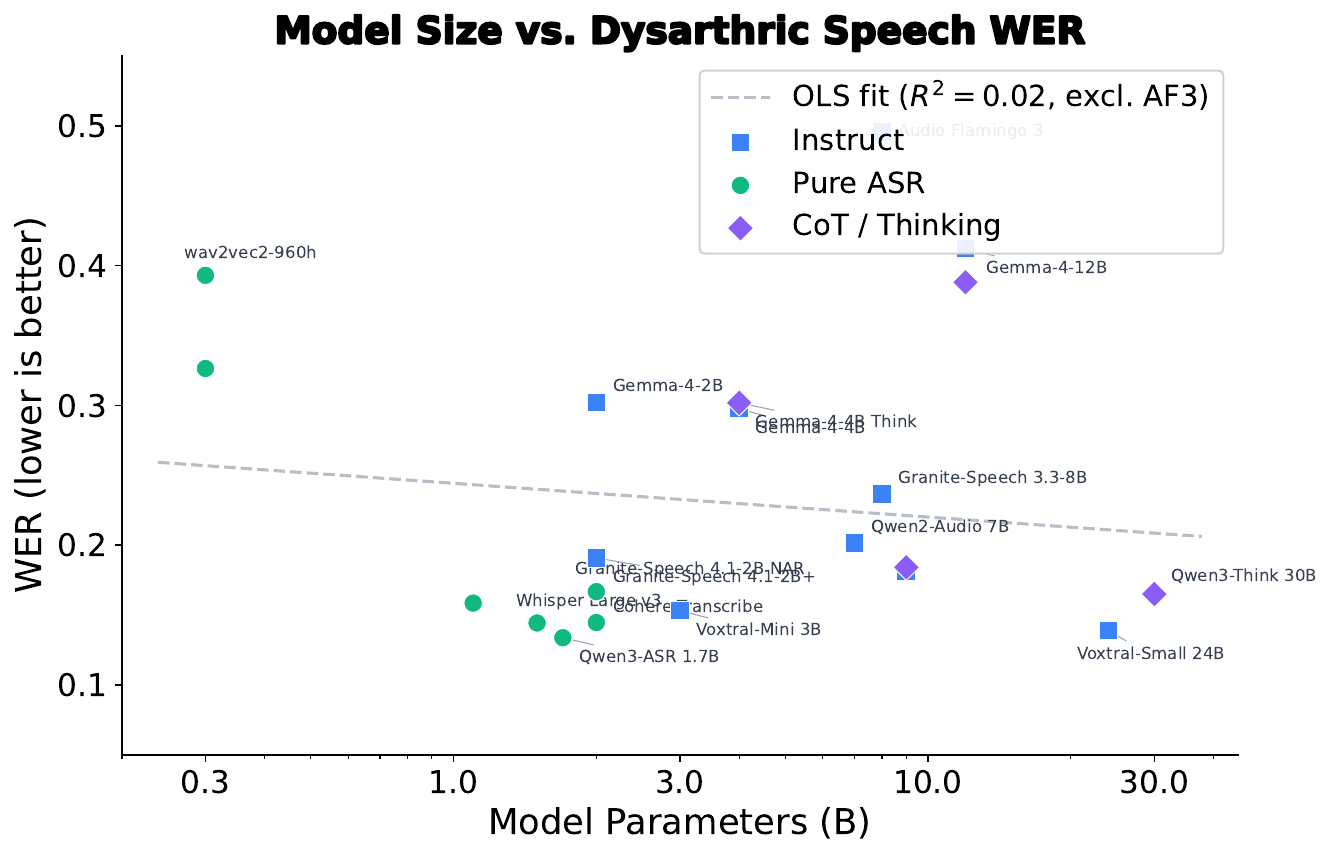}
\caption{Model size (parameters) vs.\ zero-shot WER on dysarthric speech. Dashed line shows OLS fit on log-scale parameters (excluding AF3). Model size is a weak predictor of performance ($R^2 = 0.02$); architecture and training data matter more.}
\label{fig:size_vs_wer}
\end{figure}

Figure~\ref{fig:size_vs_wer} plots model parameter count against zero-shot WER on the SAP rated subset. A log-linear OLS fit (excluding Audio Flamingo~3, whose high WER reflects output formatting rather than transcription quality; see Section~\ref{sec:results_conditioning}) yields $R^2 = 0.02$, indicating that model size alone is a weak predictor of dysarthric speech performance. Several smaller dedicated ASR models (Qwen3-ASR at 1.7B, Cohere Transcribe at 2.0B) match or outperform much larger instruct models (Qwen2-Audio at 7B, Ultravox-Nemo at 12B), suggesting that architecture and training data composition matter more than scale for this task. Notably, the three Gemma-4 variants (4B, 4B Think, and the unified 12B) cluster well above the trend line while Voxtral-Small-24B and Qwen3-Thinking-30B lie below it at larger parameter counts, further reinforcing that family and training recipe dominate over raw size.

The weak pooled fit is best understood as a confound with model class rather than a genuine size effect. Splitting by class, the correlation between $\log$ parameters and WER is strongly negative for the pure-ASR models ($r = -0.95$, $R^2 = 0.91$) but essentially absent for the instruct models ($r = -0.14$, $R^2 = 0.02$), with the chain-of-thought group too small to interpret ($N = 4$). The ASR trend, moreover, is an artifact of the two legacy Wav2Vec2 systems (0.3B, WER $>0.32$) anchoring one end against modern 1--2B ASR models at the other; it does not reflect a smooth size--accuracy relationship. The dedicated ASR models are small yet strong precisely because they are purpose-built and forgo the large language-model backbone that the instruct and CoT systems carry, whereas among those LM-backbone models WER is uncorrelated with scale (Voxtral-Small-24B at 0.139 versus Gemma-4-12B at 0.412). We therefore treat model size as globally uninformative for dysarthric ASR once model class is accounted for.

\section{Metric Breakdown and Agreement}
\label{app:metric_breakdown}
This section provides detailed metric breakdowns. Table~\ref{tab:severity_correlation} shows the correlation between ASR performance and clinical severity. Table~\ref{tab:severity_stratified} stratifies performance by severity level. Table~\ref{tab:zero_shot_results} gives the full zero-shot results with WER, CER, and SemScore by etiology. Tables~\ref{tab:prompting_wer},~\ref{tab:prompting_cer}, and~\ref{tab:prompting_semscore} break down the effect of context prompting on each metric by model and etiology, showing that WER and SemScore generally agree on the direction of prompting effects.

\begin{table}[t]
\centering
\caption{Correlation between ASR model performance and human-assessed speech severity on rated samples ($n$=11,218). Spearman $\rho$ is reported between per-sample mean severity rating (scale 1--7) and each metric. All correlations are significant at $p < 10^{-100}$. Intelligibility-specific correlation uses only samples with an explicit intelligibility dimension rating ($n$=3,738).}
\label{tab:severity_correlation}
\begin{tabular}{l ccc c}
\toprule
& \multicolumn{3}{c}{\textbf{Mean Severity} $\rho$} & \textbf{Intelligibility} \\
\cmidrule(lr){2-4} \cmidrule(lr){5-5}
\textbf{Model} & \textbf{WER} & \textbf{CER} & \textbf{SemScore} & \textbf{WER} $\rho$ \\
\midrule
Qwen3-ASR-1.7B    & .420 & .426 & $-.394$ & .494 \\
Whisper Large v3   & .401 & .412 & $-.392$ & .461 \\
Cohere Transcribe  & .380 & .392 & $-.366$ & .448 \\
Parakeet TDT 1.1B & .366 & .379 & $-.340$ & .428 \\
wav2vec2-robust    & .368 & .412 & $-.406$ & .386 \\
wav2vec2-960h      & .351 & .403 & $-.419$ & .371 \\
\bottomrule
\end{tabular}
\end{table}

\begin{table*}[t]
\centering
\caption{ASR performance stratified by mean clinical severity level (rounded). Severity 1 = normal range, 5 = most severe in the rated subset. All models are evaluated on the same 11,218 rated samples. The steepest degradation occurs between severity 2 and 3 across all models and metrics.}
\label{tab:severity_stratified}
\resizebox{\textwidth}{!}{%
\begin{tabular}{l rrrrr rrrrr rrrrr}
\toprule
& \multicolumn{5}{c}{\textbf{WER$_\mathrm{norm\_min}$ ($\downarrow$)}} & \multicolumn{5}{c}{\textbf{CER$_\mathrm{norm\_min}$ ($\downarrow$)}} & \multicolumn{5}{c}{\textbf{SemScore ($\uparrow$)}} \\
\cmidrule(lr){2-6} \cmidrule(lr){7-11} \cmidrule(lr){12-16}
\textbf{Model} & \textbf{1} & \textbf{2} & \textbf{3} & \textbf{4} & \textbf{5} & \textbf{1} & \textbf{2} & \textbf{3} & \textbf{4} & \textbf{5} & \textbf{1} & \textbf{2} & \textbf{3} & \textbf{4} & \textbf{5} \\
\midrule
Qwen3-ASR      & .044 & .113 & .357 & .477 & .736 & .028 & .076 & .261 & .358 & .657 & 87.9 & 75.0 & 49.2 & 40.6 & 40.7 \\
Cohere Trans.  & .063 & .125 & .353 & .444 & .770 & .039 & .085 & .265 & .352 & .722 & 86.1 & 73.9 & 49.7 & 41.6 & 42.0 \\
Whisper v3     & .051 & .126 & .362 & .483 & .805 & .033 & .087 & .270 & .365 & .690 & 86.7 & 73.1 & 48.0 & 39.1 & 40.5 \\
Parakeet       & .088 & .133 & .374 & .448 & .766 & .056 & .092 & .293 & .364 & .739 & 81.5 & 70.9 & 45.7 & 39.6 & 27.6 \\
w2v2-robust    & .225 & .293 & .627 & .706 & .957 & .104 & .161 & .400 & .477 & .758 & 74.8 & 61.5 & 32.4 & 24.9 & 17.2 \\
w2v2-960h      & .291 & .356 & .712 & .775 & .946 & .146 & .203 & .469 & .525 & .776 & 70.3 & 56.7 & 26.9 & 20.8 & 20.9 \\
\midrule
\textit{$n$ samples} & \textit{2,519} & \textit{7,111} & \textit{1,436} & \textit{142} & \textit{10} \\
\bottomrule
\end{tabular}%
}
\end{table*}

\begin{table*}[t]
\centering
\caption{Zero-shot ASR performance on the SAP rated subset ($N = 11{,}218$). WER and CER follow the SAP evaluation protocol (normalized, dual-reference minimum, clipped at 1.0). SemScore is the SAP composite semantic similarity metric (0--100, higher is better). All models are evaluated truly unprompted (audio only), except that Audio Flamingo~3's raw unprompted output is reported (the model produces non-transcription output without a prompt; the basic transcription instruction reduces it to 0.162). Best result per column in \textbf{bold}. $^\dagger$Mixture-of-Experts with 3B active parameters.}
\label{tab:zero_shot_results}
\resizebox{\textwidth}{!}{%
\begin{tabular}{llr rrr rrrr rrrr}
\toprule
& & & \multicolumn{3}{c}{\textbf{Overall}} & \multicolumn{4}{c}{\textbf{WER by Etiology}} & \multicolumn{4}{c}{\textbf{SemScore by Etiology}} \\
\cmidrule(lr){4-6} \cmidrule(lr){7-10} \cmidrule(lr){11-14}
\textbf{Model} & \textbf{Type} & \textbf{Params} & \textbf{WER$\downarrow$} & \textbf{CER$\downarrow$} & \textbf{Sem$\uparrow$} & \textbf{PD} & \textbf{ALS} & \textbf{DS} & \textbf{CP} & \textbf{PD} & \textbf{ALS} & \textbf{DS} & \textbf{CP} \\
\midrule
\multicolumn{14}{l}{\textit{Pure ASR Models}} \\
\midrule
Qwen3-ASR-1.7B & Pure ASR & 1.7B & \textbf{.134} & \textbf{.093} & \textbf{74.1} & .073 & \textbf{.176} & \textbf{.227} & \textbf{.333} & \textbf{79.8} & \textbf{72.5} & \textbf{59.7} & \textbf{54.4} \\
Granite-Speech 4.1-2B NAR & Pure ASR & 2B & .167 & .120 & 69.4 & .086 & .254 & .286 & .405 & 76.3 & 64.2 & 53.0 & 47.8 \\
Cohere Transcribe & Pure ASR & 2.0B & .145 & .102 & 73.1 & .076 & .191 & .289 & .365 & 79.3 & 71.4 & 54.0 & 52.3 \\
Whisper Large v3 & Pure ASR & 1.5B & .144 & .102 & 72.5 & .083 & .182 & .260 & .348 & 78.3 & 71.5 & 55.5 & 52.3 \\
Parakeet TDT 1.1B & Pure ASR & 1.1B & .159 & .114 & 69.6 & .088 & .216 & .258 & .385 & 76.0 & 67.0 & 54.4 & 48.0 \\
wav2vec2-robust & Pure ASR & 0.3B & .326 & .183 & 60.3 & .234 & .401 & .526 & .612 & 67.7 & 56.6 & 40.3 & 35.9 \\
wav2vec2-960h & Pure ASR & 0.3B & .393 & .229 & 55.4 & .294 & .486 & .598 & .688 & 63.5 & 50.6 & 34.2 & 29.7 \\
\midrule
\multicolumn{14}{l}{\textit{Instruct Audio LLMs}} \\
\midrule
Voxtral-Small 24B & Instruct & 24B & .139 & .098 & 73.3 & .075 & .183 & .257 & .347 & 79.2 & 71.8 & 56.7 & 53.0 \\
Voxtral-Mini 3B & Instruct & 3B & .153 & .106 & 72.1 & .087 & .199 & .282 & .366 & 78.1 & 70.3 & 54.2 & 51.4 \\
Phi-4 Multimodal & Instruct & 5.6B & .158 & .119 & 72.6 & .093 & .211 & .253 & .367 & 78.4 & 70.0 & 58.4 & 52.8 \\
Qwen3-Omni 30B & Instruct & 30B$^\dagger$ & .138 & .103 & 73.4 & \textbf{.068} & .194 & .239 & .365 & 79.7 & 70.7 & 58.5 & 51.8 \\
Qwen2-Audio 7B & Instruct & 7B & .201 & .153 & 68.7 & .129 & .256 & .340 & .433 & 75.3 & 65.7 & 50.7 & 46.8 \\
MiniCPM-o 4.5 & Instruct & 9B & .181 & .132 & 70.0 & .105 & .256 & .297 & .416 & 76.6 & 65.9 & 54.7 & 48.4 \\
Granite-Speech 4.1-2B+ & Instruct & 2B & .191 & .137 & 69.8 & .103 & .291 & .327 & .447 & 77.3 & 63.9 & 52.3 & 46.8 \\
Granite-Speech 3.3-8B & Instruct & 8B & .237 & .183 & 66.9 & .169 & .304 & .302 & .449 & 73.4 & 62.2 & 53.8 & 46.3 \\
Ultravox-Nemo 12B & Instruct & 12B & .312 & .244 & 54.1 & .242 & .360 & .478 & .533 & 60.1 & 50.6 & 36.6 & 35.3 \\
Gemma-4-4B & Instruct & 4B & .298 & .210 & 59.7 & .211 & .391 & .444 & .553 & 66.7 & 54.2 & 43.2 & 38.2 \\
Gemma-4-2B & Instruct & 2B & .302 & .213 & 59.5 & .216 & .397 & .479 & .550 & 66.4 & 54.0 & 41.4 & 38.7 \\
Gemma-4-12B & Instruct & 12B & .412 & .339 & 51.8 & .324 & .487 & .595 & .683 & 58.9 & 47.0 & 33.1 & 29.3 \\
Audio Flamingo 3 & Instruct & 8B & .495 & .465 & 47.7 & .428 & .557 & .701 & .686 & 52.4 & 43.9 & 35.8 & 33.8 \\
\midrule
\multicolumn{14}{l}{\textit{Chain-of-Thought}} \\
\midrule
MiniCPM-o 4.5 Think & CoT & 9B & .184 & .134 & 69.8 & .106 & .264 & .301 & .420 & 76.4 & 65.6 & 54.5 & 48.6 \\
Qwen3-Thinking 30B & CoT & 30B$^\dagger$ & .165 & .122 & 72.8 & .086 & .240 & .259 & .414 & 79.5 & 68.9 & 58.8 & 50.4 \\
Gemma-4-4B Think & CoT & 4B & .302 & .215 & 60.3 & .220 & .391 & .446 & .542 & 67.2 & 55.0 & 43.2 & 39.2 \\
Audio-Reasoner & CoT & 7B & .240 & .194 & 67.6 & .161 & .302 & .385 & .491 & 74.2 & 64.1 & 50.9 & 45.9 \\
Gemma-4-12B Think & CoT & 12B & .388 & .325 & 55.6 & .313 & .466 & .503 & .612 & 61.9 & 50.5 & 41.0 & 36.4 \\
\bottomrule
\end{tabular}%
}
\end{table*}

\begin{table*}[t]
\centering
\caption{Effect of context prompting on WER$_\mathrm{norm\_min}$ ($\downarrow$). Values show absolute P0 and $\Delta$ = P\textit{x} $-$ P0. Green indicates improvement, red degradation. For WER, negative $\Delta$ is an improvement.}
\label{tab:prompting_wer}
\resizebox{\textwidth}{!}{%
\begin{tabular}{ll rrrrr rrrrr}
\toprule
& & \multicolumn{5}{c}{\textbf{P1 (condition only)}} & \multicolumn{5}{c}{\textbf{P2 (full profile)}} \\
\cmidrule(lr){3-7} \cmidrule(lr){8-12}
\textbf{Model} & \textbf{P0} & \textbf{$\Delta$All} & \textbf{$\Delta$PD} & \textbf{$\Delta$ALS} & \textbf{$\Delta$DS} & \textbf{$\Delta$CP} & \textbf{$\Delta$All} & \textbf{$\Delta$PD} & \textbf{$\Delta$ALS} & \textbf{$\Delta$DS} & \textbf{$\Delta$CP} \\
\midrule
Audio Flamingo 3      & .498 & \imp{$-.336$}\sig{***} & \imp{$-.333$}\sig{***} & \imp{$-.346$}\sig{***} & \imp{$-.431$}\sig{***} & \imp{$-.322$}\sig{***} & \imp{$-.335$}\sig{***} & \imp{$-.331$}\sig{***} & \imp{$-.345$}\sig{***} & \imp{$-.435$}\sig{***} & \imp{$-.320$}\sig{***} \\
MiniCPM-o Think       & .184 & \imp{$-.006$} & \neu{$-.000$}\sig{***} & \imp{$-.015$}\sig{**} & \imp{$-.033$}\sig{***} & \imp{$-.016$}\sig{***} & \imp{$-.006$} & \neu{$-.000$}\sig{***} & \imp{$-.013$}\sig{*} & \imp{$-.032$}\sig{***} & \imp{$-.018$}\sig{***} \\
Qwen3-Omni 30B       & .138 & \degrad{$+.003$}\sig{***} & \degrad{$+.010$}\sig{***} & \degrad{$+.003$}\sig{***} & \imp{$-.018$}\sig{**}  & \imp{$-.020$}\sig{***} & \degrad{$+.004$}\sig{***} & \degrad{$+.010$}\sig{***} & \neu{$+.001$}\sig{**}  & \neu{$-.009$}          & \imp{$-.019$}\sig{***} \\
Phi-4 Multimodal      & .157 & \neu{$-.000$}\sig{***} & \degrad{$+.005$}\sig{***} & \neu{$-.003$}          & \imp{$-.031$}\sig{**}  & \neu{$-.015$}          & \neu{$+.000$}\sig{***} & \neu{$+.001$}\sig{***} & \degrad{$+.004$}\sig{***} & \neu{$-.002$}          & \neu{$-.007$}          \\
Voxtral-Small 24B     & .139 & \degrad{$+.010$}\sig{***} & \degrad{$+.010$}\sig{***} & \degrad{$+.014$}\sig{*}   & \neu{$+.011$}          & \neu{$+.002$}          & \degrad{$+.050$}\sig{***} & \degrad{$+.048$}\sig{***} & \degrad{$+.057$}\sig{***} & \degrad{$+.077$}\sig{***} & \degrad{$+.046$}\sig{***} \\
Qwen2-Audio 7B        & .201 & \degrad{$+.031$}\sig{***} & \degrad{$+.033$}\sig{***} & \degrad{$+.028$}\sig{***} & \degrad{$+.035$}\sig{*}   & \degrad{$+.023$}\sig{***} & \degrad{$+.051$}\sig{***} & \degrad{$+.060$}\sig{***} & \degrad{$+.029$}\sig{***} & \degrad{$+.036$}\sig{**}  & \degrad{$+.036$}\sig{***} \\
Qwen3-Thinking 30B    & .165 & \degrad{$+.006$}\sig{***} & \degrad{$+.013$}\sig{***} & \imp{$-.006$}          & \degrad{$+.005$}       & \imp{$-.012$}          & \degrad{$+.007$}\sig{***} & \degrad{$+.012$}\sig{***} & \neu{$+.001$}          & \degrad{$+.031$}\sig{*}   & \imp{$-.010$}          \\
Gemma-4-4B            & .292 & \degrad{$+.032$}\sig{***} & \degrad{$+.029$}\sig{***} & \degrad{$+.043$}\sig{***} & \degrad{$+.038$}\sig{**}  & \degrad{$+.034$}\sig{***} & \degrad{$+.049$}\sig{***} & \degrad{$+.045$}\sig{***} & \degrad{$+.068$}\sig{***} & \degrad{$+.040$}\sig{**}  & \degrad{$+.047$}\sig{***} \\
Gemma-4-4B Think      & .302 & \degrad{$+.208$}\sig{***} & \degrad{$+.225$}\sig{***} & \degrad{$+.174$}\sig{***} & \degrad{$+.178$}\sig{***} & \degrad{$+.172$}\sig{***} & \degrad{$+.202$}\sig{***} & \degrad{$+.221$}\sig{***} & \degrad{$+.175$}\sig{***} & \degrad{$+.196$}\sig{***} & \degrad{$+.147$}\sig{***} \\
Gemma-4-12B           & .412 & \degrad{$+.118$}\sig{***} & \degrad{$+.125$}\sig{***} & \degrad{$+.107$}\sig{***} & \degrad{$+.123$}\sig{***} & \degrad{$+.094$}\sig{***} & \degrad{$+.088$}\sig{***} & \degrad{$+.099$}\sig{***} & \degrad{$+.057$}\sig{***} & \degrad{$+.090$}\sig{***} & \degrad{$+.072$}\sig{***} \\
\bottomrule
\end{tabular}%
}
\end{table*}

\begin{table*}[t]
\centering
\caption{Effect of context prompting on CER$_\mathrm{norm\_min}$ ($\downarrow$). Conventions as in Table~\ref{tab:prompting_wer}.}
\label{tab:prompting_cer}
\resizebox{\textwidth}{!}{%
\begin{tabular}{ll rrrrr rrrrr}
\toprule
& & \multicolumn{5}{c}{\textbf{P1 (condition only)}} & \multicolumn{5}{c}{\textbf{P2 (full profile)}} \\
\cmidrule(lr){3-7} \cmidrule(lr){8-12}
\textbf{Model} & \textbf{P0} & \textbf{$\Delta$All} & \textbf{$\Delta$PD} & \textbf{$\Delta$ALS} & \textbf{$\Delta$DS} & \textbf{$\Delta$CP} & \textbf{$\Delta$All} & \textbf{$\Delta$PD} & \textbf{$\Delta$ALS} & \textbf{$\Delta$DS} & \textbf{$\Delta$CP} \\
\midrule
Audio Flamingo 3      & .467 & \imp{$-.349$}\sig{***} & \imp{$-.337$}\sig{***} & \imp{$-.365$}\sig{***} & \imp{$-.476$}\sig{***} & \imp{$-.363$}\sig{***} & \imp{$-.348$}\sig{***} & \imp{$-.336$}\sig{***} & \imp{$-.363$}\sig{***} & \imp{$-.478$}\sig{***} & \imp{$-.358$}\sig{***} \\
MiniCPM-o Think       & .134 & \imp{$-.004$}\sig{***} & \neu{$+.000$} & \imp{$-.010$}\sig{***} & \imp{$-.021$}\sig{***} & \imp{$-.013$}\sig{***} & \imp{$-.004$}\sig{***} & \neu{$+.000$} & \imp{$-.009$}\sig{***} & \imp{$-.017$}\sig{**} & \imp{$-.015$}\sig{***} \\
Qwen3-Omni 30B       & .103 & \imp{$-.002$}\sig{***} & \degrad{$+.005$}\sig{***} & \neu{$-.004$}          & \imp{$-.014$}\sig{*}   & \imp{$-.023$}\sig{***} & \imp{$-.001$}\sig{***} & \degrad{$+.005$}\sig{***} & \neu{$-.006$}          & \neu{$-.008$}          & \imp{$-.022$}\sig{***} \\
Phi-4 Multimodal      & .118 & \imp{$-.001$}\sig{***} & \degrad{$+.004$}\sig{***} & \neu{$-.002$}          & \imp{$-.028$}\sig{***} & \neu{$-.014$}          & \degrad{$+.001$}\sig{***} & \neu{$+.000$}\sig{***} & \degrad{$+.007$}\sig{***} & \neu{$+.001$}          & \neu{$-.002$}          \\
Voxtral-Small 24B     & .098 & \degrad{$+.011$}\sig{***} & \degrad{$+.009$}\sig{***} & \degrad{$+.020$}\sig{***} & \degrad{$+.021$}\sig{*}   & \degrad{$+.011$}\sig{*}   & \degrad{$+.054$}\sig{***} & \degrad{$+.047$}\sig{***} & \degrad{$+.068$}\sig{***} & \degrad{$+.093$}\sig{***} & \degrad{$+.062$}\sig{***} \\
Qwen2-Audio 7B        & .153 & \degrad{$+.016$}\sig{***} & \degrad{$+.020$}\sig{***} & \degrad{$+.010$}\sig{***} & \neu{$+.008$}          & \neu{$+.002$}          & \degrad{$+.037$}\sig{***} & \degrad{$+.046$}\sig{***} & \degrad{$+.017$}\sig{***} & \neu{$+.018$}          & \degrad{$+.022$}\sig{***} \\
Qwen3-Thinking 30B    & .122 & \degrad{$+.004$}\sig{***} & \degrad{$+.010$}\sig{***} & \imp{$-.002$}          & \neu{$+.004$}          & \imp{$-.013$}          & \degrad{$+.006$}\sig{***} & \degrad{$+.010$}\sig{***} & \neu{$+.003$}          & \degrad{$+.029$}\sig{*}   & \imp{$-.012$}          \\
Gemma-4-4B            & .205 & \degrad{$+.028$}\sig{***} & \degrad{$+.026$}\sig{***} & \degrad{$+.036$}\sig{***} & \degrad{$+.034$}\sig{**}  & \degrad{$+.031$}\sig{***} & \degrad{$+.050$}\sig{***} & \degrad{$+.045$}\sig{***} & \degrad{$+.076$}\sig{***} & \degrad{$+.046$}\sig{**}  & \degrad{$+.052$}\sig{***} \\
Gemma-4-4B Think      & .215 & \degrad{$+.226$}\sig{***} & \degrad{$+.234$}\sig{***} & \degrad{$+.203$}\sig{***} & \degrad{$+.221$}\sig{***} & \degrad{$+.216$}\sig{***} & \degrad{$+.223$}\sig{***} & \degrad{$+.233$}\sig{***} & \degrad{$+.204$}\sig{***} & \degrad{$+.250$}\sig{***} & \degrad{$+.190$}\sig{***} \\
Gemma-4-12B           & .339 & \degrad{$+.145$}\sig{***} & \degrad{$+.148$}\sig{***} & \degrad{$+.135$}\sig{***} & \degrad{$+.168$}\sig{***} & \degrad{$+.138$}\sig{***} & \degrad{$+.116$}\sig{***} & \degrad{$+.122$}\sig{***} & \degrad{$+.085$}\sig{***} & \degrad{$+.137$}\sig{***} & \degrad{$+.116$}\sig{***} \\
\bottomrule
\end{tabular}%
}
\end{table*}

\begin{table*}[t]
\centering
\caption{Effect of context prompting on SemScore ($\uparrow$, scale 0--100). Conventions as in Table~\ref{tab:prompting_wer}. For SemScore, positive $\Delta$ is an improvement.}
\label{tab:prompting_semscore}
\resizebox{\textwidth}{!}{%
\begin{tabular}{ll rrrrr rrrrr}
\toprule
& & \multicolumn{5}{c}{\textbf{P1 (condition only)}} & \multicolumn{5}{c}{\textbf{P2 (full profile)}} \\
\cmidrule(lr){3-7} \cmidrule(lr){8-12}
\textbf{Model} & \textbf{P0} & \textbf{$\Delta$All} & \textbf{$\Delta$PD} & \textbf{$\Delta$ALS} & \textbf{$\Delta$DS} & \textbf{$\Delta$CP} & \textbf{$\Delta$All} & \textbf{$\Delta$PD} & \textbf{$\Delta$ALS} & \textbf{$\Delta$DS} & \textbf{$\Delta$CP} \\
\midrule
Audio Flamingo 3      & 51.1 & \imp{$+18.3$}\sig{***} & \imp{$+19.2$}\sig{***} & \imp{$+20.5$}\sig{***} & \imp{$+15.0$}\sig{***} & \imp{$+13.3$}\sig{***} & \imp{$+18.3$}\sig{***} & \imp{$+19.1$}\sig{***} & \imp{$+20.4$}\sig{***} & \imp{$+15.6$}\sig{***} & \imp{$+13.1$}\sig{***} \\
MiniCPM-o Think       & 69.8 & \neu{$+0.6$} & \neu{$+0.2$}\sig{***} & \neu{$+1.2$} & \imp{$+2.7$}\sig{***} & \imp{$+1.5$}\sig{***} & \imp{$+0.9$}\sig{***} & \neu{$+0.5$} & \imp{$+1.5$}\sig{*} & \imp{$+2.8$}\sig{***} & \imp{$+1.6$}\sig{***} \\
Qwen3-Omni 30B       & 73.4 & \imp{$+0.6$}\sig{***}  & \neu{$+0.1$}\sig{***}  & \imp{$+0.9$}\sig{***}  & \imp{$+2.1$}\sig{***}  & \imp{$+2.3$}\sig{***}  & \imp{$+0.6$}\sig{***}  & \neu{$+0.1$}\sig{***}  & \imp{$+1.0$}\sig{***}  & \imp{$+1.3$}\sig{***}  & \imp{$+2.1$}\sig{***}  \\
Phi-4 Multimodal      & 72.6 & \degrad{$-0.5$}\sig{***}  & \degrad{$-1.0$}\sig{***}  & \neu{$+0.3$}\sig{**}   & \neu{$+2.4$}           & \neu{$+0.4$}           & \degrad{$-0.5$}\sig{***}  & \degrad{$-0.6$}\sig{***}  & \neu{$-0.1$}\sig{*}    & \neu{$+0.2$}           & \neu{$-0.5$}           \\
Voxtral-Small 24B     & 73.1 & \neu{$-0.2$}           & \neu{$-0.2$}\sig{*}    & \neu{$-0.4$}           & \neu{$-0.3$}           & \neu{$+0.1$}\sig{***}  & \degrad{$-3.6$}\sig{***}  & \degrad{$-3.3$}\sig{***}  & \degrad{$-4.6$}\sig{***}  & \degrad{$-5.6$}\sig{***}  & \degrad{$-3.5$}\sig{***}  \\
Qwen2-Audio 7B        & 68.7 & \degrad{$-3.9$}\sig{***}  & \degrad{$-4.5$}\sig{***}  & \degrad{$-3.1$}\sig{***}  & \degrad{$-2.6$}\sig{***}  & \degrad{$-2.3$}\sig{***}  & \degrad{$-5.3$}\sig{***}  & \degrad{$-6.5$}\sig{***}  & \degrad{$-2.9$}\sig{***}  & \degrad{$-2.6$}\sig{*}    & \degrad{$-2.9$}\sig{***}  \\
Qwen3-Thinking 30B    & 72.8 & \neu{$-0.4$}\sig{***}     & \neu{$-0.9$}\sig{***}     & \neu{$+0.2$}              & \neu{$+0.0$}              & \imp{$+1.0$}\sig{***}     & \neu{$-0.5$}\sig{***}     & \neu{$-0.9$}\sig{***}     & \neu{$+0.1$}              & \degrad{$-1.7$}\sig{**}   & \neu{$+0.5$}              \\
Gemma-4-4B            & 60.9 & \degrad{$-2.8$}\sig{***}  & \degrad{$-2.7$}\sig{***}  & \degrad{$-3.3$}\sig{***}  & \degrad{$-3.1$}\sig{**}   & \degrad{$-3.0$}\sig{***}  & \degrad{$-4.6$}\sig{***}  & \degrad{$-4.3$}\sig{***}  & \degrad{$-6.5$}\sig{***}  & \degrad{$-3.2$}\sig{**}   & \degrad{$-4.4$}\sig{***}  \\
Gemma-4-4B Think      & 60.3 & \degrad{$-15.6$}\sig{***} & \degrad{$-16.4$}\sig{***} & \degrad{$-15.1$}\sig{***} & \degrad{$-12.5$}\sig{***} & \degrad{$-12.8$}\sig{***} & \degrad{$-13.5$}\sig{***} & \degrad{$-14.9$}\sig{***} & \degrad{$-12.5$}\sig{***} & \degrad{$-11.9$}\sig{***} & \degrad{$-9.2$}\sig{***}  \\
Gemma-4-12B           & 51.8 & \degrad{$-10.4$}\sig{***} & \degrad{$-11.0$}\sig{***} & \degrad{$-10.2$}\sig{***} & \degrad{$-9.0$}\sig{***}  & \degrad{$-8.2$}\sig{***}  & \degrad{$-8.3$}\sig{***}  & \degrad{$-9.0$}\sig{***}  & \degrad{$-6.5$}\sig{***}  & \degrad{$-7.1$}\sig{***}  & \degrad{$-7.2$}\sig{***}  \\
\bottomrule
\end{tabular}%
}
\end{table*}

\section{Additional Analyses}
\label{app:additional}
This appendix collects analyses summarised in the main text: the iterative-correction condition, the deployment profile of context-dependent fine-tuning, the word-level error decomposition, and supporting figures and subgroup tables.

\subsection{Hallucination and Verbosity: Full Analysis}

\begin{figure}[tbp]
\centering
\includegraphics[width=\textwidth]{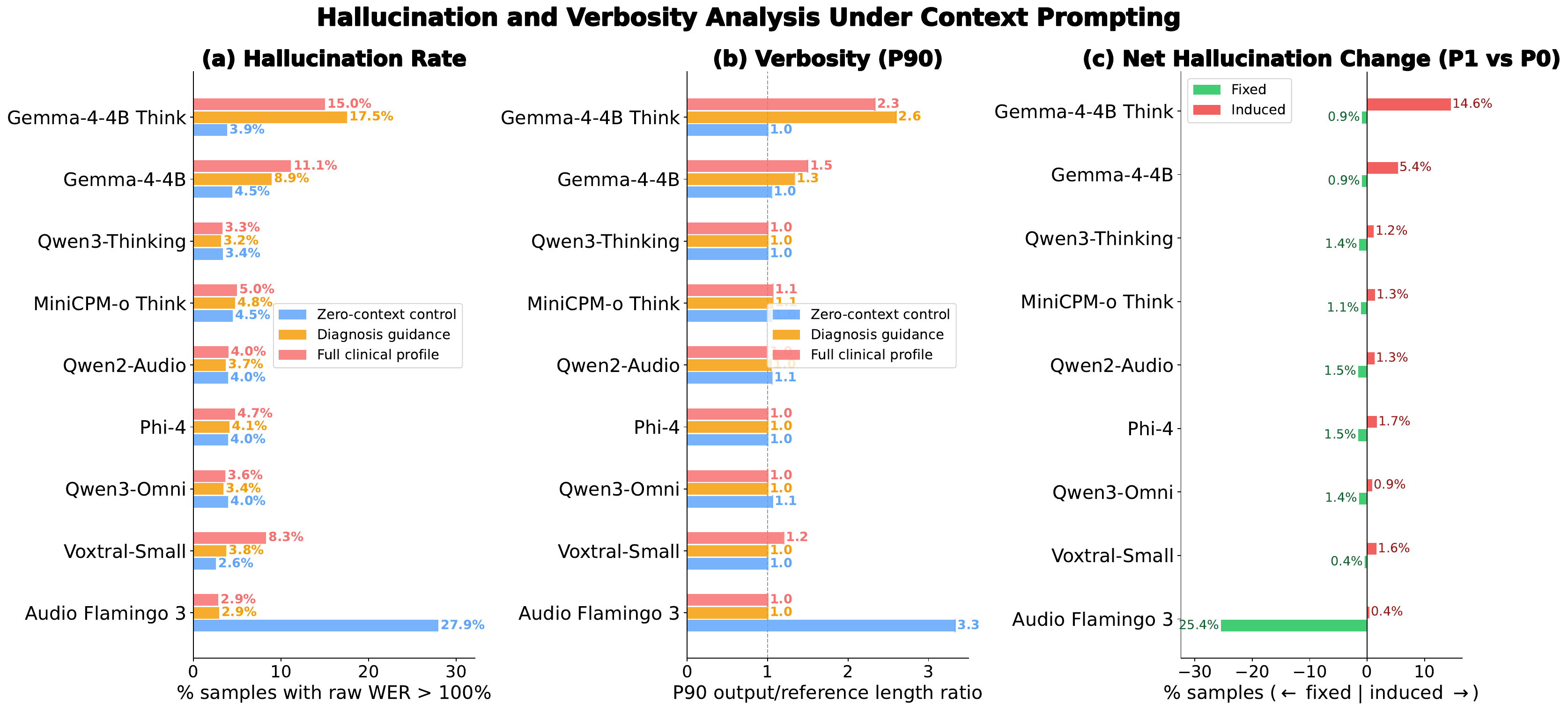}
\caption{Hallucination and verbosity under context prompting ($N = 11{,}218$). (a)~Hallucination rate (raw WER $> 100$\%). (b)~P90 output-to-reference length ratio; dashed line marks parity. (c)~Samples fixed vs.\ induced by diagnosis guidance.}
\label{fig:hallucination_failure_modes}
\end{figure}

A persistent concern with context prompting is whether clinical information simultaneously increases catastrophic failures. We analyze hallucination rates, verbosity, and failure modes across all frozen models under zero-context (P0), diagnosis-only (P1), and full clinical profile (P2) conditions (Figure~\ref{fig:hallucination_failure_modes}, Appendix~\ref{app:additional}).

We define a hallucination as any sample where unclipped WER exceeds 100\%, indicating the prediction contains more erroneous words than the reference. Under zero-context decoding, hallucination rates range from 2.6\% (Voxtral-Small) to 27.9\% (Audio Flamingo~3), with the latter driven by a systematic preamble artifact (\texttt{"The spoken content of the audio is~'...'"}) that inflates word counts. Context prompting has divergent effects. For Audio Flamingo~3, diagnosis guidance reduces the hallucination rate from 27.9\% to 2.9\%, nearly eliminating the preamble mode. Voxtral-Small, by contrast, increases from 2.6\% to 8.3\% under P2 due to verbose reformulations triggered by the clinical profile. Gemma-4-4B Think shows the most extreme case, with the hallucination rate increasing from 3.9\% at zero-context to 17.5\% under diagnosis guidance and 15.0\% under the full clinical profile. Qwen3-Thinking, by contrast, remains stable across conditions (3.3--3.4\%), and the remaining models also show $\pm$1\,pp variation.

The 90th percentile of output-to-reference length ratios (Figure~\ref{fig:hallucination_failure_modes}b) confirms that most models maintain a P90 near 1.0 regardless of condition. Audio Flamingo~3 under P0 is the exception at 3.3$\times$, normalizing to 1.0 once prompting constrains the output format, and Gemma-4-4B Think inflates from 1.0 at zero-context to 2.3--2.6$\times$ under prompting, consistent with its catastrophic-degradation behaviour.

Decomposing the rate change into samples \textit{fixed} (hallucinating under P0 but not P1) versus \textit{induced} (not hallucinating under P0 but hallucinating under P1), Audio Flamingo~3 fixes approximately 25\% of all samples (Figure~\ref{fig:hallucination_failure_modes}c). Gemma-4-4B Think shows the opposite pattern: diagnosis guidance induces hallucinations in 14.6\% of samples while fixing almost none. Among the remaining models, induced and fixed rates are small and roughly balanced (1--3\% each), indicating that context prompting does not systematically create new failure modes in well-behaved instruct models.

Manual inspection reveals five recurring failure patterns. (1)~\textit{Repetitive loops}, where the model echoes disfluency patterns described in the clinical profile. (2)~\textit{Verbose reformulation}, where the model paraphrases rather than transcribes. (3)~\textit{Language switching}, producing fragments in German or Chinese when confronted with severely distorted speech. (4)~\textit{Refusal} (``I'm sorry, I cannot\ldots''). (5)~\textit{Reasoning chain leakage}, unique to CoT models, where internal deliberation text appears in the output. Of these, only~(1) is directly attributable to the clinical context. The model appears to ``anticipate'' described disfluency patterns and inadvertently reproduce them.

\begin{figure}[tbp]
\centering
\includegraphics[width=0.45\textwidth]{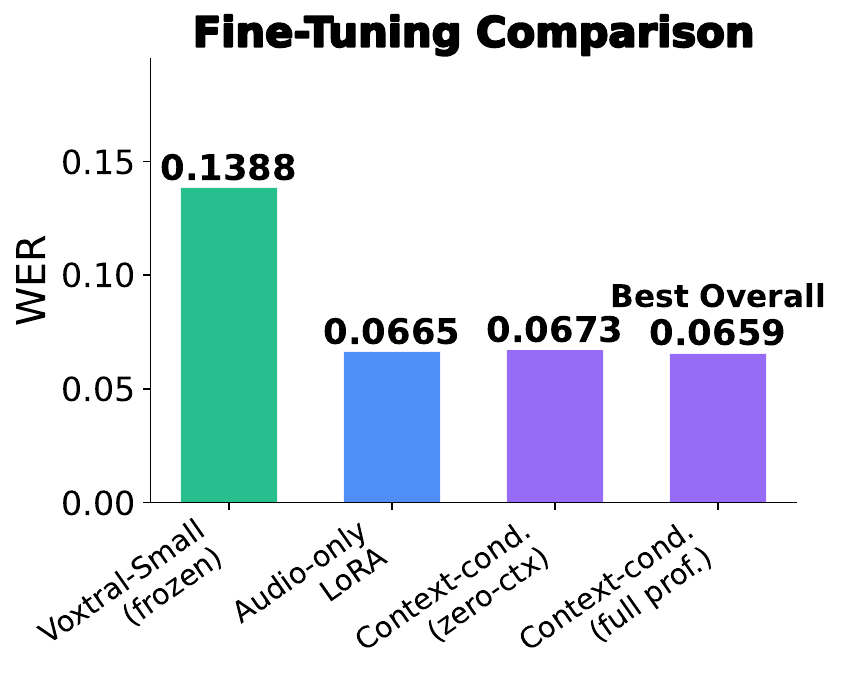}
\caption{Fine-tuning comparison (5-fold cross-validation, 437 speakers). The context-dependent system achieves the best overall WER. Numerical values are in Table~\ref{tab:supervised_summary}.}
\label{fig:supervised_results_appendix}
\end{figure}

\subsection{Iterative Correction}
\label{sec:results_iterative}
The follow-up correction condition (Table~\ref{tab:conditions} in Appendix~\ref{app:prompts}) tests whether models can iteratively improve a transcription when given both the full clinical profile and a prior-pass transcript. This two-pass setup represents the prompting scenario where the model receives the audio, complete clinical context, and an initial transcription attempt to refine.

Across models where this condition was evaluated, follow-up correction does not systematically improve WER over single-pass prompting. Comparing follow-up correction to the full clinical profile in Table~\ref{tab:conditioning}, WER is effectively unchanged for AF3 ($-$0.0002), Phi-4 ($+0.0020$), and Qwen3-Omni ($+0.0001$); Voxtral-Small ($-$0.0165) recovers some ground from the prompt-induced verbose output seen under the full profile; Qwen2-Audio shows no such recovery and instead worsens further ($+$0.0048 vs.\ full profile), with both models still underperforming the zero-context baseline by $+$0.0336 and $+$0.0558 WER respectively. The prior transcript does not systematically improve recognition, consistent with the broader finding that frozen models do not meaningfully condition on text-based context when processing audio, and suggests that iterative probing alone is unlikely to overcome this limitation.

\subsection{Context-dependent Fine-Tuning in Practice}
The context-dependent fine-tuning results (Sections~\ref{sec:results_supervised} and~\ref{sec:results_subgroups}) show that clinical context becomes useful once the model has been trained to expect it. The subgroup pattern (largest gains for Down syndrome and mild severity, no gain for moderate, slight degradation for cerebral palsy) is consistent with a model that benefits from context where the audio-only system struggles. The difficulty analysis (Section~\ref{sec:results_difficulty}) makes this precise: the benefit grows with baseline difficulty and peaks on the hardest utterances, while clearly transcribable speech gains nothing and can pay a small penalty.

From a deployment standpoint, the approach has a favourable risk profile. When clinical context is unavailable, the context-dependent model performs comparably to the audio-only baseline ($p = 0.93$). When context is available, it helps. This asymmetry, with no penalty for missing context and modest benefit when present, makes context-dependent fine-tuning a strategy for clinical settings where metadata availability varies across patients and sessions. The difficulty-aware gate sharpens this further: rather than applying context indiscriminately, a deployed system should invoke it selectively on utterances where the audio-only pass is likely failing (Section~\ref{sec:results_difficulty}).

\subsection{Word-Level Error Decomposition}
\label{sec:wordlevel}
To characterize what the context-dependent system repairs, we align the audio-only and context-dependent predictions word by word against the reference, type each error by CMU-dictionary phonological class (consonant, vowel, word-boundary, other), and measure how often the context-dependent system recovers a reference word the audio-only system missed (speaker-level statistics, $N = 11{,}218$ utterances).

\begin{figure}[tbp]
\centering
\includegraphics[width=\textwidth]{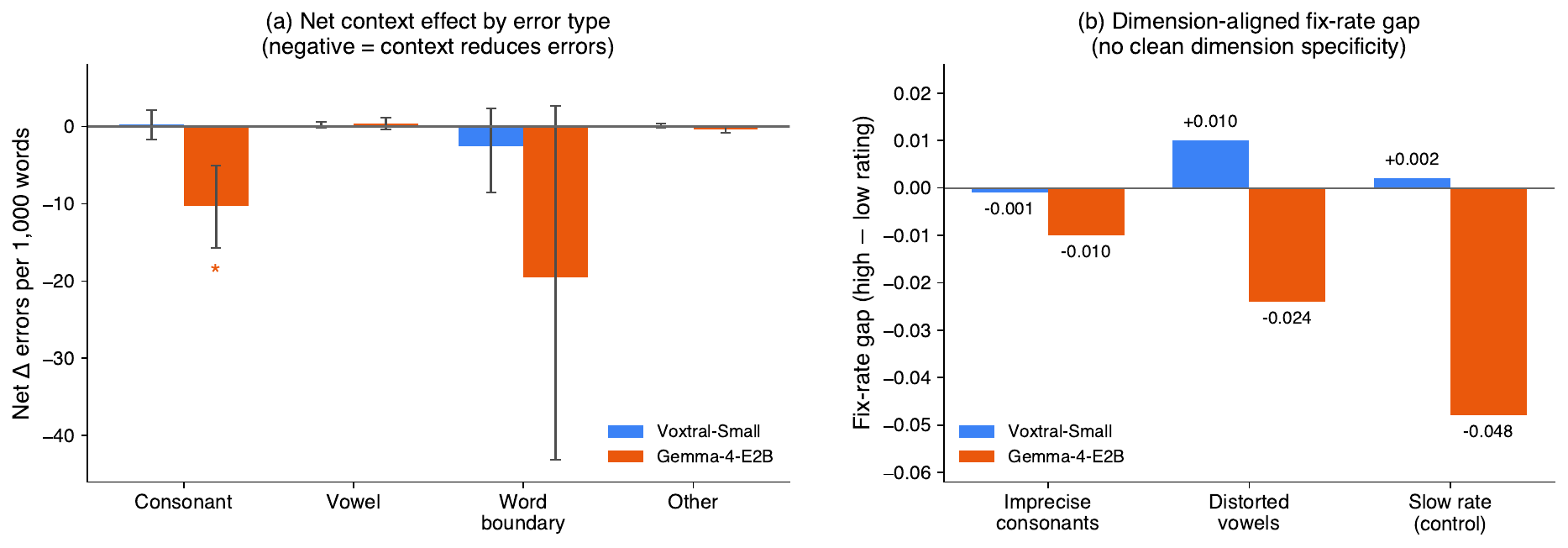}
\caption{Word-level error decomposition of the context-dependent vs.\ audio-only LoRA systems (5-fold cross-validation, $N = 11{,}218$). (a)~Net change in errors per 1{,}000 reference words from adding the full clinical profile (negative = context reduces that error type), with bootstrap intervals. On Voxtral-Small all four types are null; on Gemma-4-E2B consonant errors drop significantly ($\ast$, interval excluding zero). (b)~Fix-rate gap between samples rated high vs.\ low on each clinical dimension. The dimension-aligned gaps are small and the prosodic control (slow rate) is comparable, so the residual recovery is not directed by the prompt's content.}
\label{fig:wordlevel}
\end{figure}

Context-dependent fine-tuning does repair pronunciation errors: the context-dependent system recovers 17\% of the audio-only system's consonant-word errors and 20\% of its vowel-word errors. The net effect, however, is architecture-dependent (Figure~\ref{fig:wordlevel}a). On Voxtral-Small the recoveries are offset by new errors, and the net change per 1{,}000 reference words is null for all four error types (every interval crossing zero). On Gemma-4-E2B, context significantly reduces consonant errors (about 10 per 1{,}000 words, interval excluding zero), consistent with the weaker model using context to repair consonant-heavy speech.

The repair is not specific to the matching clinical dimension (Figure~\ref{fig:wordlevel}b). If the model extracted graded information, samples rated high on \emph{imprecise consonants} should show a larger consonant fix-rate gap than samples rated low, and analogously for \emph{distorted vowels}. The dimension-aligned gaps are negligible on Voxtral-Small ($-0.001$ for imprecise consonants, $+0.010$ for distorted vowels, $+0.002$ for the prosodic control) and show no clean pattern on Gemma-4-E2B either ($-0.010$ and $-0.024$, with the slow-rate control at $-0.048$). A counterfactual cross-check settles the mechanism: when the clinical profile is replaced with a different speaker's, the high-versus-low gap by the sample's \emph{true} rating persists rather than disappearing. The recovery therefore tracks the acoustic difficulty of the sample, not the content of the prompt. Our reading is that the fine-tuned model uses clinical context as a coarse signal that triggers lexical recovery on degraded audio, concentrated where the audio-only model is failing, rather than as graded per-dimension information, consistent with the difficulty-conditioned benefit of Section~\ref{sec:results_difficulty} and with the hallucinated-correction risk discussed in Section~\ref{sec:conclusion}.

As a further probe we also ran the chain-of-thought (\emph{Think}) variant of the unified 12B on P0--P2 (Tables~\ref{tab:baselines},~\ref{tab:paired_headline}). Enabling thinking \emph{lowers} the zero-context WER ($0.388$ vs.\ $0.412$) and suppresses the refusals almost entirely (from $\approx$$24\%$ to under $2\%$), but it does not rescue the prompted condition: the profile-induced degradation is if anything slightly larger ($\Delta$WER $+0.113$ vs.\ $+0.088$), converging to the same $\approx$$0.50$ WER. This within-family contrast is itself informative, for the smaller Gemma-4-4B, thinking instead \emph{amplifies} the degradation (Table~\ref{tab:paired_headline}), and points to future mechanistic analysis of when a reasoning trace aids versus displaces the transcription.

\begin{table}[tbp]
\caption{Subgroup analysis: audio-only vs.\ context-dependent. Rel.\% is the aggregate relative WER change, $(\overline{\mathrm{WER}}_{\mathrm{ctx}} - \overline{\mathrm{WER}}_{\mathrm{a0}}) / \overline{\mathrm{WER}}_{\mathrm{a0}}$ (negative = improvement). $p$ from Wilcoxon signed-rank at speaker level.}
\label{tab:subgroups_main}
\vspace{2pt}
\small
\centering
\begin{minipage}[t]{0.48\columnwidth}
\centering
\textbf{By Etiology}\\[2pt]
\begin{tabular}{lrrrc}
\toprule
 & $N$ & Rel.\% & $p$ \\
\midrule
Down syndrome  & 358   & $-$7.1 & \textbf{.044} \\
ALS            & 1{,}594 & $-$4.6 & .066 \\
Parkinson's    & 7{,}504 & $-$2.7 & .231 \\
Cerebral palsy & 1{,}762 & +3.1   & .121 \\
\bottomrule
\end{tabular}
\end{minipage}%
\hfill
\begin{minipage}[t]{0.48\columnwidth}
\centering
\textbf{By Severity}\\[2pt]
\begin{tabular}{lrrrc}
\toprule
 & $N$ & Rel.\% & $p$ \\
\midrule
Mild (1--2)     & 7{,}304 & $-$2.7 & \textbf{.033} \\
Moderate (3--4) & 3{,}893 & +0.0   & .900 \\
Severe (5--7)   & 21      & $-$1.8 & .750 \\
\bottomrule
\end{tabular}
\end{minipage}
\end{table}

\begin{figure}[tbp]
\centering
\includegraphics[width=\columnwidth]{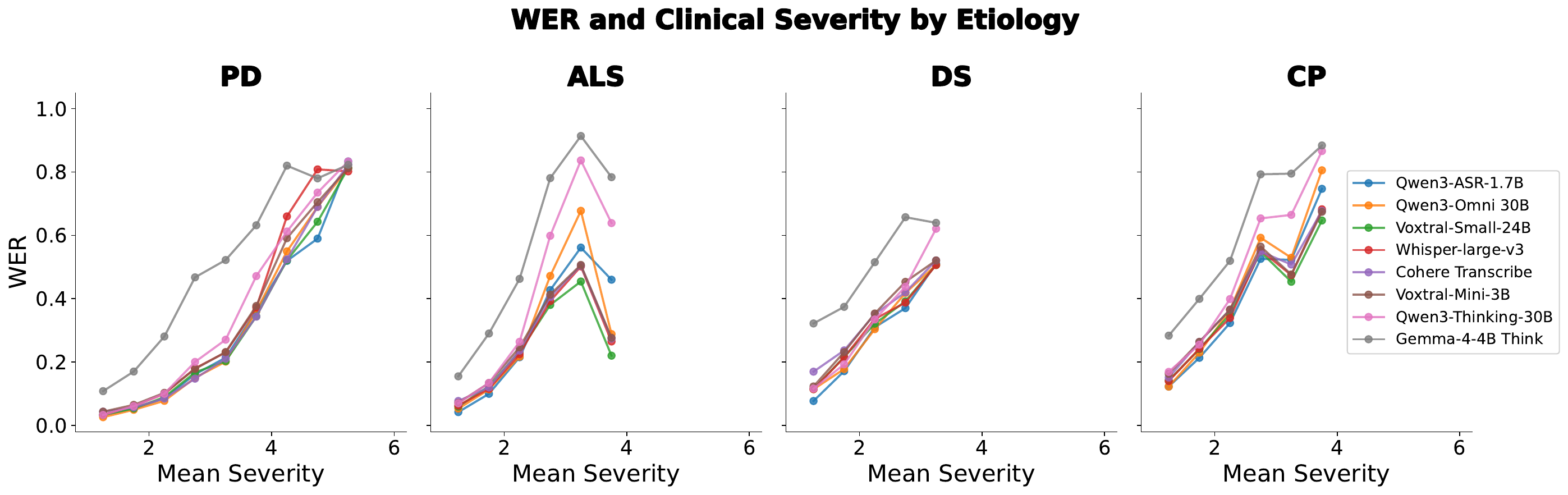}
\caption{WER vs.\ clinical severity by etiology. Performance degrades beyond severity level 2 across models, with the steepest decline for cerebral palsy and Down syndrome.}
\label{fig:severity_wer}
\end{figure}

\begin{figure}[tbp]
\centering
\includegraphics[width=\textwidth]{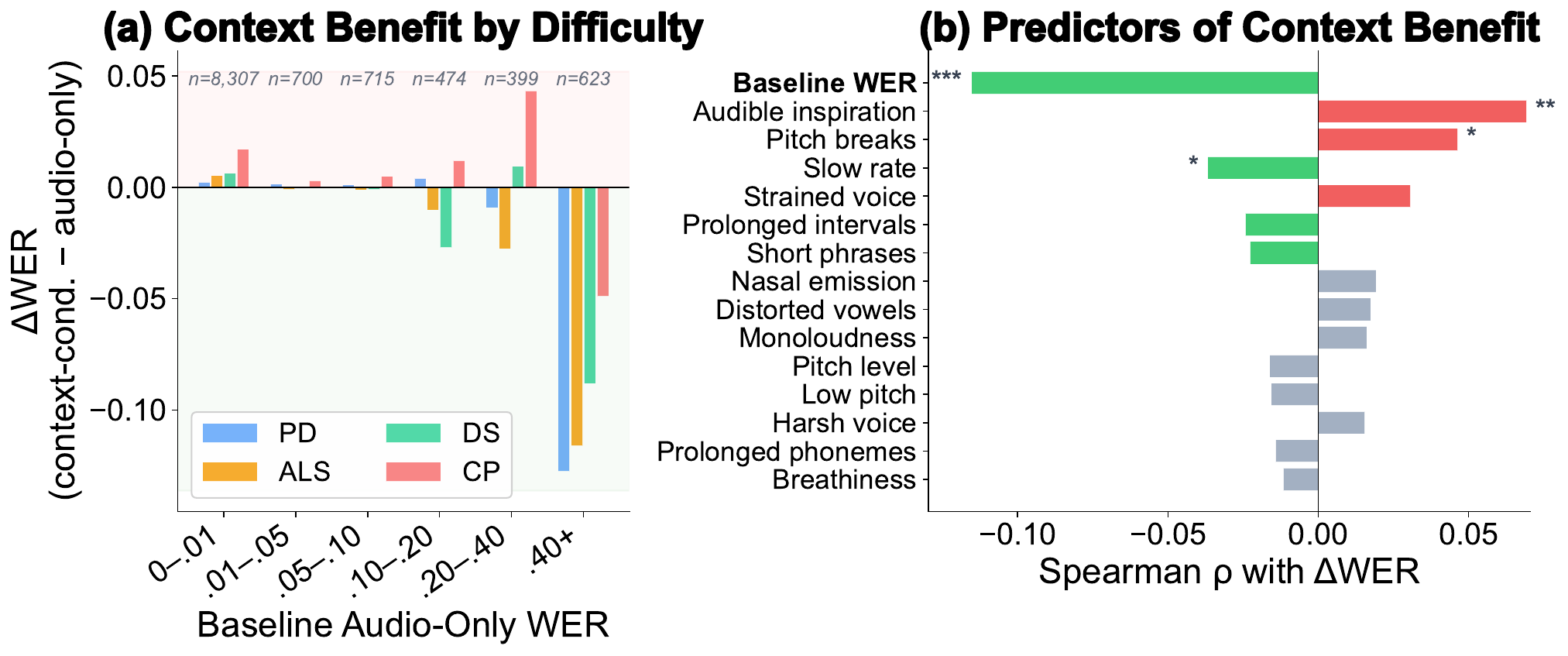}
\caption{Predictors of per-sample context benefit (audio-only $\to$ context-dependent, 5-fold cross-validation, $N = 11{,}218$). (a)~Mean $\Delta$WER by baseline difficulty and etiology. (b)~Spearman $\rho$ between each predictor and $\Delta$WER. Baseline WER dominates; individual clinical dimensions show near-zero correlation.}
\label{fig:context_benefit_predictors_appendix}
\end{figure}

\end{document}